\documentclass[10pt,twocolumn,letterpaper]{article}
\usepackage{cvpr}

\usepackage{amsmath,amsfonts,bm}

\def\eqref#1{equation~\ref{#1}}
\def\Eqref#1{Equation~\ref{#1}}

\def\1{\bm{1}}

\DeclareMathAlphabet{\mathsfit}{\encodingdefault}{\sfdefault}{m}{sl}
\SetMathAlphabet{\mathsfit}{bold}{\encodingdefault}{\sfdefault}{bx}{n}

\usepackage[accsupp]{axessibility}

\usepackage{times}
\usepackage[utf8]{inputenc} %
\usepackage[T1]{fontenc}    %
\usepackage{url}            %
\usepackage{amsfonts,amsmath,amssymb}       %
\usepackage{nicefrac}       %
\usepackage{multirow}
\usepackage{algorithm,algorithmic}
\usepackage[normalem]{ulem}
\usepackage{arydshln}
\usepackage{caption}
\usepackage{enumitem}
\usepackage{wrapfig}
\usepackage{microtype}
\usepackage{graphicx}
\usepackage{booktabs} %
\usepackage{cutwin}
\usepackage[pangram]{blindtext}
\usepackage[calc]{adjustbox}
\usepackage{subcaption}
\usepackage{color}
\usepackage{svg}

\usepackage{amsthm}

\definecolor{Red}{rgb}{0.6,0,0}
\definecolor{Blue}{rgb}{0,0,0.8}
\definecolor{Green}{rgb}{0,0.4,0.7}
\definecolor{airforceblue}{rgb}{0.36, 0.54, 0.66}
\definecolor{ao(english)}{rgb}{0.0, 0.5, 0.0}
\definecolor{azure(colorwheel)}{rgb}{0.0, 0.5, 1.0}
\definecolor{crimson}{rgb}{0.86, 0.08, 0.24}
\definecolor{darkcerulean}{rgb}{0.03, 0.27, 0.49}
\definecolor{cobalt}{rgb}{0.0, 0.28, 0.67}
\definecolor{rosegold}{rgb}{0.72, 0.43, 0.47}
\definecolor{orange-red}{rgb}{1.0, 0.27, 0.0}
\definecolor{mountainmeadow}{rgb}{0.19, 0.73, 0.56}
\definecolor{malachite}{rgb}{0.04, 0.85, 0.32}
\definecolor{darkblue}{rgb}{0.0, 0.0, 0.55}

\definecolor{color1}{HTML}{5CA45C}
\definecolor{color2}{HTML}{F6AB1F}
\definecolor{color3}{HTML}{10539A}
\definecolor{color4}{HTML}{B0182D}
\definecolor{color5}{HTML}{8685bf}
 
\definecolor{redfigure}{RGB}{240, 87, 87}
\definecolor{bluefigure}{RGB}{64,105,225}
\definecolor{greenfigure}{RGB}{34,178,170}

\usepackage[backref=page]{hyperref}
\hypersetup{colorlinks=true}
\hypersetup{linktoc=all}
\hypersetup{citecolor=MidnightBlue}
\hypersetup{linkcolor=Red}
\hypersetup{urlcolor=MidnightBlue}
\usepackage[all]{hypcap}
\usepackage[noabbrev, nameinlink, capitalize]{cleveref}

\usepackage{xcolor,colortbl}
\definecolor{nvgreen}{RGB}{118, 185, 0}

\definecolor{lgreen}{RGB}{245,245,245}

\usepackage{hyperref}
\hypersetup{colorlinks, citecolor={teal}, urlcolor=black, urlbordercolor=black}

\usepackage[square,numbers]{natbib}
\title{Heterogeneous Continual Learning}
\def\cvprPaperID{9801} %
\def\confName{CVPR}
\def\confYear{2023}

\author{
Divyam Madaan$^{1,2}\thanks{Work done during an internship at NVIDIA.}$ \ , 
Hongxu Yin$^{1}$,  
Wonmin Byeon$^{1}$, 
Jan Kautz$^{1}$,
Pavlo Molchanov$^{1}$ \\
\vspace{-0.35cm}
\\
$^1$NVIDIA, $^2$New York University \\
\tt\small divyam.madaan@nyu.edu, 
\tt\small \{dannyy, wbyeon, jkautz, pmolchanov\}@nvidia.com
}

\begin{document}
\maketitle

\begin{abstract}

We propose a novel framework and a solution to tackle the continual learning (CL) problem with
\textit{changing network architectures}. Most CL methods focus on adapting a single architecture to a new task/class by modifying its weights. However, with rapid progress in architecture design, the problem of adapting existing solutions to novel architectures becomes relevant. To address this limitation, we propose \emph{Heterogeneous Continual Learning (HCL)}, where a wide range of evolving network architectures emerge continually together with novel data/tasks. As a solution, we build on top of the distillation family of techniques and modify it to a new setting where a weaker model takes the role of a teacher; meanwhile, a new stronger architecture acts as a student. Furthermore, we consider a setup of limited access to previous data and propose \emph{Quick Deep Inversion (QDI)} to recover prior task visual features to support knowledge transfer. QDI significantly reduces computational costs compared to previous solutions and improves overall performance. In summary, we propose a new setup for CL with a modified knowledge distillation paradigm and design a quick data inversion method to enhance distillation. Our evaluation of various benchmarks shows a significant improvement on accuracy in comparison to state-of-the-art methods over various networks architectures.

\end{abstract}
\section{Introduction}
Over the past decade, we have seen a plethora of innovations in deep neural networks (DNNs) that have led to remarkable improvement in performance on several applications~\citep{ramesh2021zero,mehta2021delight, mcgrath2021acquisition}. AlexNet~\citep{krizhevsky12nips} was the first breakthrough showing the potential of deep learning on the ImageNet benchmark~\citep{deng2009imagenet}, it was followed by various architectural advances such as VGG~\citep{simonyan2014very}, Inception~\citep{szegedy2017inception}, ResNet and its variants~\citep{he2016deep,zagoruyko2016wide,xie2017aggregated}, efficient architectures~\citep{howard2017mobilenets, tan2019efficientnet}, and applications of Transformer~\citep{vaswani2017attention} in computer vision~\citep{dosovitskiy2021an, liu2021swin}; however, all these architectures are trained from scratch and compared on ImageNet classification~\citep{deng2009imagenet}. In real-world scenarios, the dataset size is not constant, yet storage of billions of data instances and retraining the model representations from scratch is computationally expensive and infeasible. Further, access to old datasets is often not available due to data privacy. Thus, it is crucial to transfer the knowledge learned by the previous model representations to the recent state-of-the-advances without forgetting knowledge and accessing prior datasets.

\begin{figure*}[t!]
\centering
\resizebox{.95\linewidth}{!}{%
\includegraphics[]{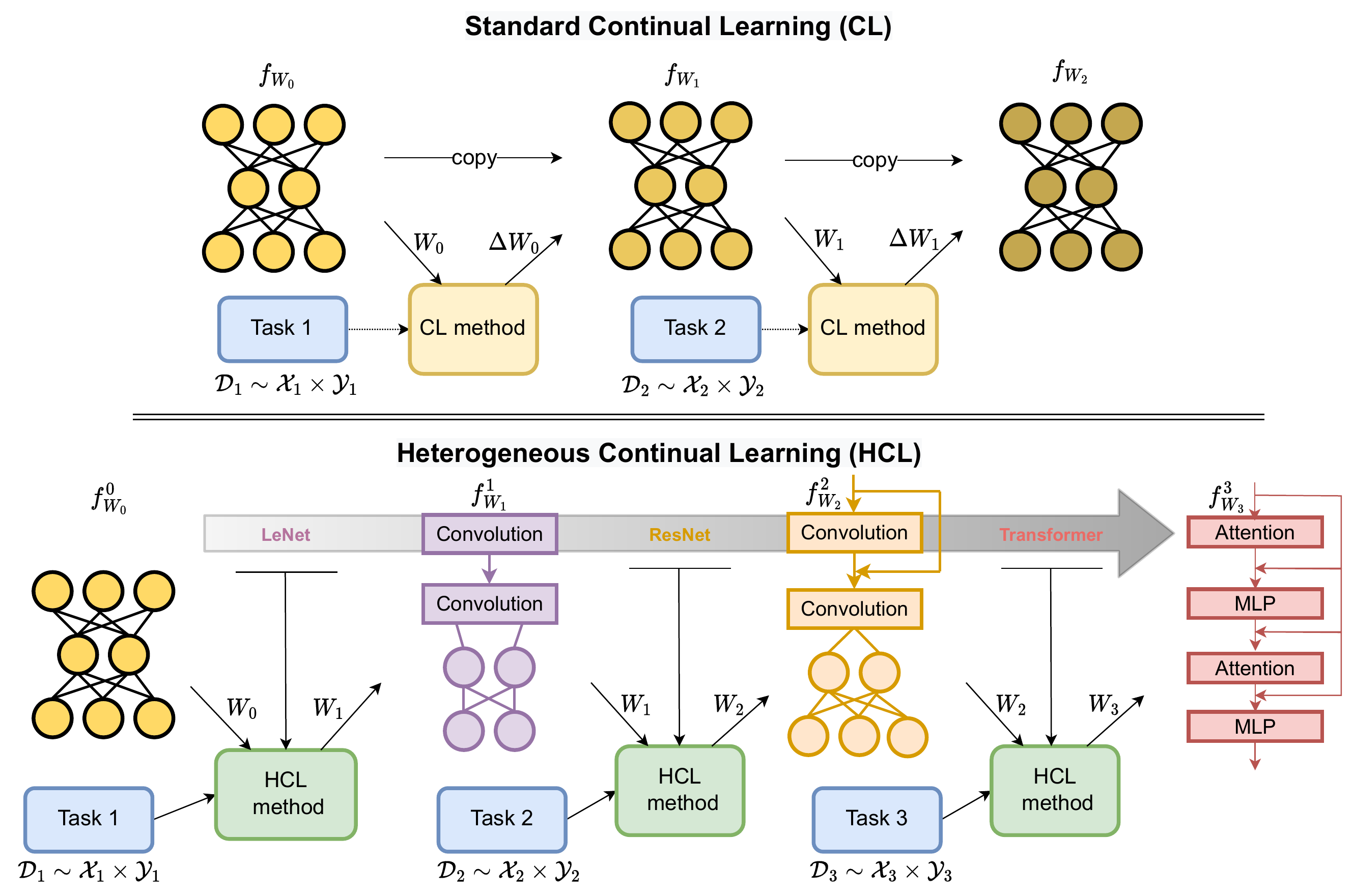}
}
\caption{\textbf{Illustration of standard and heterogenous continual learning.} Standard CL updates the same representational structure while preserving previous information with incoming tasks. On the contrary, the goal of HCL is to continuously evolve the network architecture while maintaining the accuracy of new data instances.
\label{fig:concept}}
\vspace{-0.1in}
\end{figure*}

Continual learning (CL)~\citep{thrun1996learning} is a common way to train representations on a data stream without forgetting the learned knowledge. 
However, all prior CL techniques~\citep{zenke17si, ahn2019uncertainty, rusu2016progressive,rebuffi2017icarl, rolnick19er, buzzega2020dark, madaan2022representational} continually adapt a fixed representation structure -- a fully-connected network with two hidden layers for MNIST~\citep{lecun1998mnist} and standard ResNet-18~\citep{he2016deep} for all the other datasets (see \Cref{fig:concept}). While recent works investigate the effect of depth and width~\citep{mirzadeh2022wide} on continual learning, to our knowledge, no work has focused on the real problem of learning on sequentially arriving tasks with changing network architectures. Moreover, most works assume the weight transfer with the same architecture~\citep{rusu2016progressive, YoonJ2018iclr}, in case of the previous model being available only as black box, these methods are not applicable.

In this work, we argue that maintaining the same network structure during training is not a realistic assumption for practical applications such as autonomous driving, clinical applications, and recommendation systems. Instead, constant upgrades to stronger models with the best-known architecture are vital for the best customer experience and competitive advantage, and all streams of previous data are not only large but difficult to store, transfer, manage, and protect against security breach.
As a motivating example, consider the tasks of clinical diagnosis and segmentation using medical images, where it is essential to continually update the model on arriving data to adapt to the changing environment. However, continually adapting the old model architectures can potentially hurt the current and future task performance. Also, it is often not allowed to retain the previous patients' data to update to the new model due to privacy concerns. 
Based on this motivation, our goal is to continually update the state-of-the-art deep learning architectures without storing the previous data while preserving the knowledge from previously learned representations.

We propose a novel CL framework called \emph{Heterogeneous Continual Learning (HCL)} to learn continual representations on a data stream without any restrictions on the training configuration or network architecture of the representations (see \Cref{fig:concept}). In particular, we consider the continual learner as a stream of learners, where we train each learner with the same or different backbone architecture. To this end, we consider a diverse range of heterogeneous architectures including LeNet~\citep{lecun1998gradient}, ResNet and its variants~\citep{he2016deep,zagoruyko2016wide,xie2017aggregated}, and recent architectures such as RegNet~\citep{xu2022regnet} and vision transformers~\citep{liu2021swin}. However, due to the assumption of the same architectural structure across different tasks, conventional regularization~\citep{LiZ2016learning, zenke17si, schwarz2018progress, ahn2019uncertainty} and architectural methods~\citep{rusu2016progressive, YoonJ2018iclr, li2019learn} are not directly applicable for HCL. 

To continually learn heterogeneous representations, we revisit knowledge distillation~\citep{hinton2015distilling} -- a method to transfer the knowledge learned across different networks. While knowledge distillation has been formulated for continual learning~\citep{LiZ2016learning, rebuffi2017icarl, buzzega2020dark}, prior methods do not outperform the experience replay based methods~\citep{rebuffi2017icarl, rolnick19er, Aljundi2019GradientBS, buzzega2020dark}. Moreover, the distillation pipeline has not been considered for network evolution. This work shows that knowledge distillation can outperform state-of-the-art methods with and without replay examples in standard and heterogeneous continual learning settings with proper modifications. Specifically, we incorporate \emph{consistency} and \emph{augmentation}~\citep{beyer2022knowledge} into the knowledge distillation paradigm for CL and low-temperature transfer with label-smoothing~\citep{chandrasegaran2022revisiting} for learning the network representations. Our solution comes at the cost of storing and reusing only the most recent model.

The continual learner might not have access to the prior data. Therefore, motivated by DeepInversion (DI)~\citep{yin2020dreaming}, we consider \emph{data-free continual learning}. In particular, DI optimizes random noise to generate class-conditioned samples; however, it requires many steps to optimize a single batch of instances. This increases its computational cost and slows CL. In response to this, we propose \emph{Quick Deep Inversion (QDI)}, which utilizes current task examples as the starting point for synthetic examples and optimizes them to minimize the prediction error on the previous tasks. Specifically, this leads to an interpolation between the current task and previous task instances, which promotes current task adaptation while minimizing catastrophic forgetting. We compare our proposed method on various CL benchmarks against state-of-the-art methods, where it outperforms them across all settings with a wide variety of architectures.
In summary, the contributions of our work are:
\begin{itemize}
    \item We propose a novel CL framework called \emph{Heterogeneous Continual Learning (HCL)} to learn a stream of different architectures on a sequence of tasks while transferring the knowledge from past representations. 
\item We revisit knowledge distillation and propose \emph{Quick Deep Inversion (QDI)}, which inverts the previous task parameters while interpolating the current task examples with minimal additional cost.
    \item We benchmark existing state-of-the-art solutions in the new setting and outperform them with our proposed method across a diverse stream of architectures for both task-incremental and class-incremental CL.
\end{itemize}

\section{Related work} 
\noindent{\bf Replay-based approaches.} These approaches formulate different criterion~\citep{rebuffi2017icarl, rolnick19er, Aljundi2019GradientBS, buzzega2020dark} to select a representative subset that is revisited in future tasks. \textcolor{black}{While most approaches~\citep{rebuffi2017icarl, rolnick19er, Aljundi2019GradientBS} store the input examples, \citet{buzzega2020dark} retain the logits to preserve the knowledge from the past.} However, all these methods are limited in practical scenarios due to privacy concerns and excessive memory requirements to store the data from previous tasks. Our approach does not strictly require a buffer to store data or representation, and we also do not assume that the learners’ architecture is fixed.

\noindent{\bf Regularization methods.} These methods penalize the change in the representations by adding a regularization term in the loss~\citep{LiZ2016learning, zenke17si, schwarz2018progress, ahn2019uncertainty}. \citet{LiZ2016learning} minimizes the change in the output representations, and \citet{zenke17si} estimates the important parameters and restricts their change during training. However, the regularization cannot be enforced when the architectures are different across tasks.

\noindent{\bf Architectural methods.} They consider the inclusion of task-dependent representations to prevent catastrophic forgetting during continual learning~\citep{rusu2016progressive, YoonJ2018iclr, li2019learn, dai2020incremental, yoon2020apd}. \citet{rusu2016progressive} instantiates a new column for the new task and require inputs from both columns and \citet{yoon2020apd} decomposes each layer parameters to task-specific and task-shared parameters. The objective of these methods is to reduce forgetting and the network expansion is restricted to the same structure. In contrast, we do not make any assumption on the network structure and incorporate the recent advances in deep learning to train the continual learner without any knowledge of prior data or its architecture. To the best of our knowledge, this is the first work that explores continual learning with different representational structures while adapting to the current task and minimizing forgetting. 

\noindent{\bf Optimal architectures for continual learning.} Prior works also investigate most optimal architecture for continual learning. \citet{lee21sharing} formulated a data-driven layer-based transfer function to learn the layers that should be transferred to mitigate forgetting. \citet{mirzadeh2021wide} showed that increasing the width mitigates forgetting, whereas increasing the depth negatively affects forgetting, and \citet{mirzadeh2022architecture} investigated the effect of various components in an architecture. \citet{carta2022ex} proposed Ex-Model Continual Learning to learn the continual learner from a stream of trained models; 
however, their proposed scheme is not directly applicable to the HCL setting as the architecture of their learner is fixed and it requires an extra buffer to store the past expert and CL representations.

\section{Heterogeneous Continual Learning}
In real-world scenarios, the data distribution continuously evolves with time; thus, adapting the network representations to the distribution shifts is essential while preserving the knowledge on the past sequence of data. Meanwhile, deep learning techniques are improving rapidly, providing better model architectures and techniques with higher performance on various tasks.
However, prior CL methods~\citep{zenke17si, ahn2019uncertainty, rusu2016progressive,rebuffi2017icarl, rolnick19er, buzzega2020dark} either optimize the performance on a single architecture~\citep{he2016deep} or focus on finding the optimal architecture for CL~\citep{mirzadeh2021wide, mirzadeh2022architecture}.

\begin{wrapfigure}{rt!}{0.5\linewidth}
 \vspace{-0.1in} %
\hspace{-0.36in}
    \centering
    \begin{minipage}[t]{1.05\linewidth}
        \includegraphics[width=\linewidth]{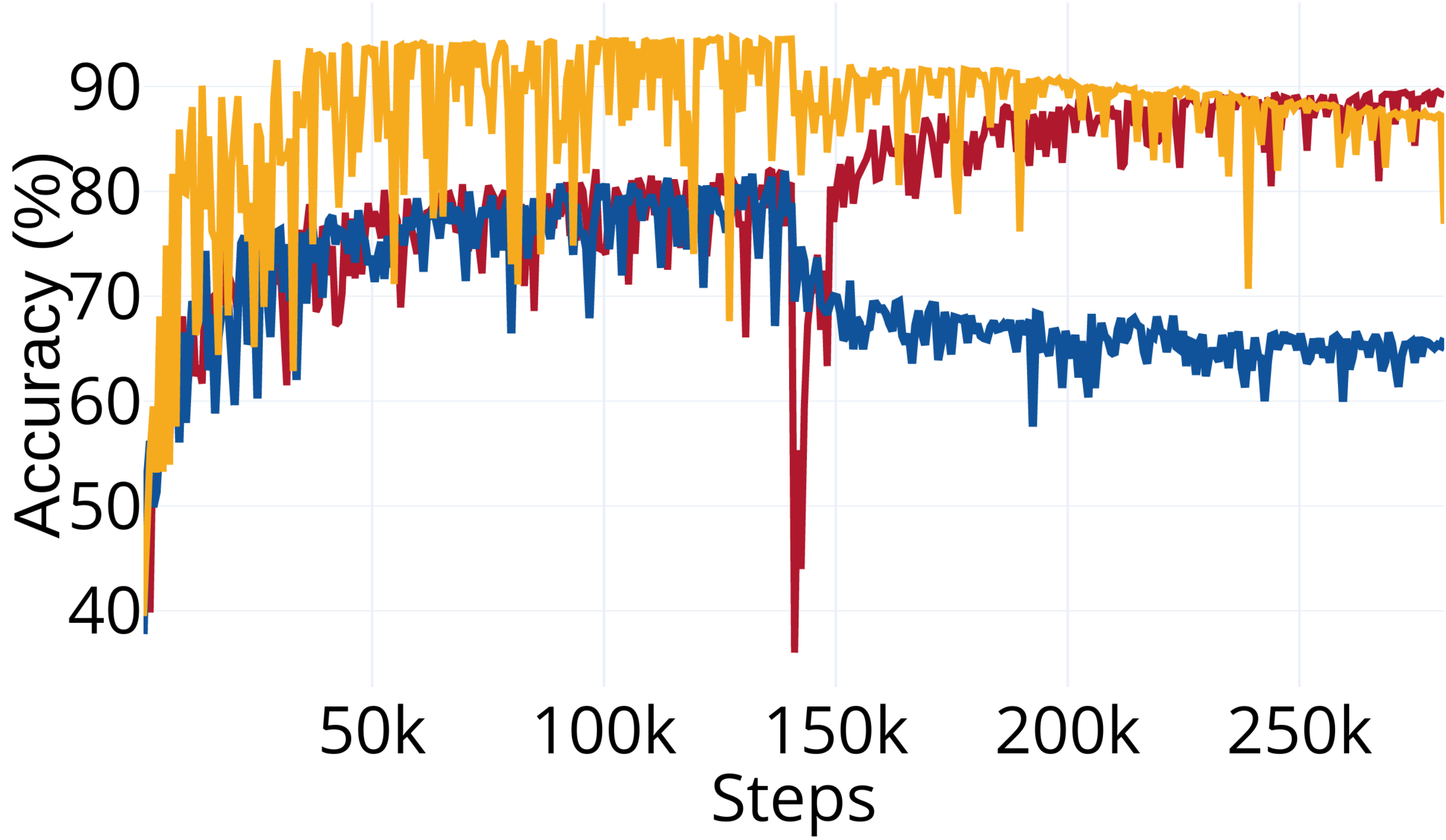}
          \caption{\small Average accuracy comparison between 
           standard CL -- \textcolor{color2}{ResNet18} and \textcolor{color3}{LeNet} trained with buffer, \textcolor{color4}{ours} with adaptive architectures without buffer on CIFAR-10 dataset with two tasks containing five classes each. 
           \label{fig:motivation}}
    \end{minipage}
          \vspace{-0.12in}
\end{wrapfigure}

To motivate further, we provide a simple experiment on CIFAR-10 split into two tasks demonstrating the benefit of adapting to novel neural architectures in \Cref{fig:motivation}. With the new ability to switch the network architecture, 
(\textit{e.g.,} from LeNet to ResNet) the performance surpasses standard CL setup that is limited by a rigid but weak architecture (\textcolor{color4}{HCL} vs. \textcolor{color3}{CL LeNet}). 
More appealingly, it even approximates the stronger model deployed throughout the CL process (\textcolor{color4}{HCL} vs. \textcolor{color2}{CL ResNet}), while alleviating the requirement of replay buffer to store prior data experiences (\textcolor{color4}{HCL is data-free} vs. \textcolor{color3}{CL with buffer}), as we will show next in methodology and experiment sections.

\subsection{Problem setup}
The problem of standard CL considers the learning objective where the continual learner $f_{W_t}: \mathcal{X}_t \rightarrow \mathcal{Y}_t$ learns a sequence of $T$ tasks by updating the {\bf fixed representation structure} for each task. Each task $t\in T$ contains training data $\mathcal{D}_t=\{x_{i}, y_{i}\}_{i=1}^{N_t} \sim \mathcal{X}_t \times \mathcal{Y}_t$, which is composed of $N_t$ identically and independently distributed examples. More formally, we minimize the following objective:
\begin{equation}
    {\underset{W_t}{\text{minimize}}}~~ \mathbb{E}_{(x_i, y_i)\sim \mathcal{D}_t}[\ell\left(f_{W_t}(x_i), y_i \right)],
\end{equation}
where $\ell: \mathcal{Y}_t \times \mathcal{Y}_t \rightarrow \mathbb{R}_{\geq0}$ is the task-specific loss function. In this work, we consider the continual learner as a {\bf stream of architectures} $\{f^1_{W_1},\ldots, f^t_{W_t}\}$, where the learner can completely change the backbone architecture to incorporate recent architectural developments~\citep{he2016deep,dosovitskiy2021an, liu2021swin, tolstikhin2021mlpmixer} in order to improve model performance. 

However, when the architectures are different, there is no natural knowledge transfer mechanism, and the network parameters are initialized randomly, which makes this problem challenging compared to standard continual learning.

Particularly, each architectural representation \(f^t_{W_t}~:~\mathcal{X}_t \rightarrow \mathcal{Y}_t,~ \forall {t \in \{1,\ldots,T\}}\) is trained on task distribution \(\mathcal{D}_t\) and our objective is to train these stream of networks on a sequence of tasks without forgetting the knowledge of the previous set of tasks. Additionally, when the network structure remains the same, we want to transfer the learned representations sequentially to train incoming tasks.
Overall, the learning objective of the HCL is:
\begin{equation}
    {\underset{W_t}{\text{minimize}}}~~ \mathbb{E}_{(x_i, y_i)\sim \mathcal{D}_t}[\ell\left(f^t_{W_t}(x_i), y_i \right)], 
\end{equation}
For notation simplicity, we discard $W_t$ in $f^t_{W_t}$ for the rest of the paper. In this work, we focus on the continual learner that does not rely on task identifiers during training and uses constant memory following prior works~\citep{buzzega2020dark, de2021continual}. We consider task-incremental (task-IL) and class-incremental learning (class-IL) during inference but do not use task labels during training. 
In task-IL, we provide the task identity to select the classification head, whereas class-IL uses a shared head across all classes. Consequently, class-IL is more challenging due to an equal weight for all the tasks and prone to higher forgetting and lower accuracy than task-IL; however, recent works~\citep{cossu2022class} also raise concerns regarding the practicality of class-IL in real-world scenarios.

\subsection{Proposed method}
\begin{figure}
  \begin{center}
    \includegraphics[width=0.8\linewidth]{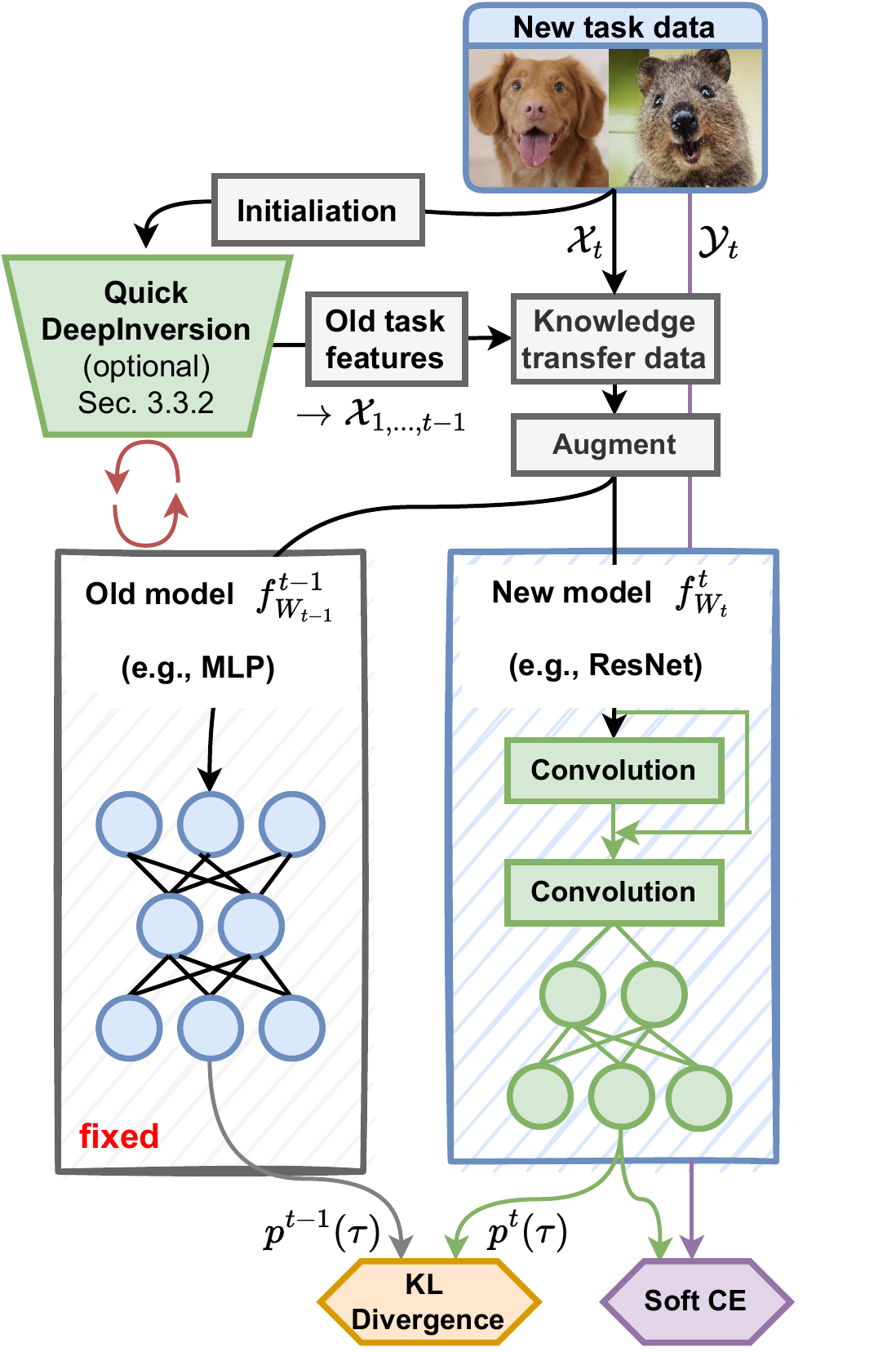}
  \end{center}
  \vspace{-0.1in}
    \caption{\small Proposed method to perform continual learning with neural network architecture change.\label{fig:method}}
 
  \vspace{-0.1in}
\end{figure}

We address the problem of HCL with our method summarized in~\Cref{fig:method}. As in the standard CL, new task data and an optional buffer are used to directly train a new model (stronger student) initialized from scratch with a task loss.
Additionally, a knowledge transfer data (KDA) is collected given new task data,  (optional) buffer and (optional) synthesised samples by quick deep inversion.
The KDA supports network transfer via knowledge distillation between old and new models. To describe our method, we first revisit knowledge distillation~\citep{hinton2015distilling} in \Cref{subsec:kd} and propose a training paradigm that outperforms state-of-the-art methods. Then, we introduce quick deep inversion in \Cref{subsec:qdi} to recover past visual features promptly. It inverts the previous task model to generate synthetic examples

\subsubsection{Revisiting knowledge distillation}\label{subsec:kd}

Due to a difference in model representations and the absence of the training data used to train previous models, we concentrate on knowledge distillation (KD)~\citep{hinton2015distilling} for HCL. However, our goal is orthogonal to standard KD; instead of distilling the knowledge from a large model (or ensemble) to a smaller (weaker) model for efficient deployment we distill from the latter to the former. Our motivation is to improve model performance by transferring to a state-of-the-art model.

While KD has been adopted in CL~\citep{LiZ2016learning,rebuffi2017icarl, Hou19learning, dhar2019learning, buzzega2020dark}, it is (i) limited to homogeneous architectures across different tasks, and (ii) considered a weak baseline in comparison to replay-based methods. Contrarily, we focus on HCL and propose a modified paradigm inspired by the recent advances in KD to improve the performance and lift both limitations. 

\noindent{\bf Label smoothing.} Prior works~\citep{muller2019does, shen2021is} investigating the compatibility between label smoothing (LS) and knowledge distillation have derived opposite conclusions. Recently, \citet{ chandrasegaran2022revisiting} highlighted that LS-trained teacher with a low-temperature transfer is essential to achieve students with higher performance. This inspired us to adapt label smoothing to produce better soft targets for KD, which improves the knowledge transfer from a smaller to a larger model architecture reducing the forgetting for both the standard CL and HCL scenarios. 

\noindent{\bf Knowledge distillation.} To improve the KD paradigm for HCL, we incorporate \emph{consistency} and \emph{augmentation}~\citep{beyer2022knowledge}. Specifically, fixed teaching and distillation without augmentation leads to overfitting of the larger model on the current task performance. On the contrary, consistent teaching using the same input views for the student and teacher and augmentation improves generalization while reducing forgetting.
In contrast to standard knowledge distillation, we use the current task instances for distillation, as the previous task data is often unavailable in real-world applications due to data privacy and legal restrictions. 

Additionally, our method is applicable to task-free continual learning during training as it does not require task identity. In addition, we do not use warmup before the current task as we find that it is not essential to improve model performance. We conduct ablation experiments to justify the design choices of every component in \Cref{tab:l2_u}. Our overall objective for the lifelong learner is:
\begin{equation}\label{eq:our_objective}
\small 
    {\underset{W_t}{\text{minimize}}}~~ \mathbb{E}_{(x_i, y_i)\sim \mathcal{D}_t}[\ell\left(f^t(x_i),  y^t_i(\psi)\right)] + \alpha \cdot \text{KL}\left(p^{t}_i(\tau), p^{t-1}_i(\tau)\right),
\end{equation}
where $y^t_i(\psi) = y (1 - \psi) + \psi/C$, $\psi$ denotes the mixture parameter to interpolate the hard targets to uniform distribution defined using $\psi$, $C$ is the number of classes, $p^t(\tau)$ and $p^{t-1}(\tau)$ denote the temperature-scaled output probabilities for current task model and past-task model, respectively, with $\tau$ as the corresponding temperature. $\alpha$ is the hyper-parameter that controls the strength of KD loss.

\subsubsection{Enhancing data efficiency}\label{subsec:qdi}

To further improve our data-free HCL paradigm, we extend DeepInversion (DI)~\citep{yin2020dreaming}, which optimizes images $\tilde{x}$ initialized with Gaussian noise to improve knowledge transfer from the prior tasks given only the trained models and no data. The objective of the optimization is to excite particular features, or, as in our work, classes from previous tasks.  
The optimization is guided by proxy image priors: (i) total variation $\mathcal{L}_{tv}$, (ii) $\mathcal{L}_{\ell_2}$ of the generated samples and (iii) feature distribution regularization $\mathcal{L}_{\text{feature}}$. DI requires many steps to generate examples prior to every task, which increases the computational cost of CL method and, therefore, becomes less appealing.

To tackle this limitation, we propose \emph{Quick Deep Inversion (QDI)} that initializes the synthetic examples with 
the current task data prior to the optimization. There is no reason to believe that randomly sampled Gaussian noise is an optimal starting point for generating past task examples. We optimize the current task examples such that features will fall to the manifold learned by the previous model and the domain shift is minimized. As a result such imags are classified as past task classes.  
QDI generates examples \({\tilde{x}}_{\text{prior}} \{f^{t-1}, k\}\) to approximate features from all prior tasks $\{1,\ldots,t-1\}$ by inverting the last model $f^{t-1}$ with $k$ optimization steps:
\begin{align}\label{eq:di_objective}
   {\tilde{x}}_{\text{prior}} \{f^{t-1}, k\} =  \underset{\tilde{x}}{\text{argmin}}~~\mathcal{L}(f^{t-1}(\tilde{x}), \tilde{y}) \nonumber \\ + \alpha_{tv}\mathcal{L}_{tv}(\tilde{x}) + \alpha_{\ell_2} \mathcal{L}_{\ell_2}(\tilde{x}) + \alpha_{\text{feature}} \mathcal{L}_{\text{feature}},
\end{align}
where the synthesized examples are optimized towards prior classes $\tilde{y} \sim \mathcal{Y}_{\{1, ..., t-1\}}$
to minimize forgetting. 
$\alpha_{tv}, \alpha_{\ell_2}, \alpha_{\text{feature}}$ denote the hyper-parameters that determine the strength of individual losses. 
To improve synthesis speed, we initialize $\tilde{x}$ with the current task input image $x_t$, which provides a $4 \times$ speed-up and leads to more natural images:
\begin{align}
    \tilde{x}_{\text{prior}} \{f^{t-1}, 0\} = \tilde{x}_{\text{prior}, \: k=0} = x_t. \nonumber
\end{align}

The initialized $\tilde{x}_{\text{prior}}$ can then optimized using Equation~\ref{eq:di_objective}, regularized for realism by the target model $f^{t-1}$, 
hence quickly unveiling previous task visual features (see \Cref{fig:di_qdi}) on top of the current task image through 
\begin{align}\label{eq:di_objective}
\small
    \mathcal{L}_{\text{feature}} = \sum_{l\in L} \Big[d\Big(\mu_l(\tilde{x}_{\text{prior}}), \mathbb{E}\big[\mu_l(x) | x \sim {\mathcal{X}_{\{1, ..., t-1\}}}\big] \Big) \nonumber \\ + \: d\Big(\sigma_l(\tilde{x}_{\text{prior}}), \mathbb{E}\big[\sigma_l(x) | x \sim {\mathcal{X}_{\{1, ..., t-1\}}}\big] \Big)  \Big].  &
\end{align}
$d(\cdot, \cdot)$ denotes the distance metric for feature regularization. In this paper, we use the mean-squared distance. Akin to~\citet{yin2020dreaming}, we assume Gaussian distribution for feature maps and focus on batch-wise mean \(\mu_l(x)\) and variance \(\sigma_l(x)\) for the layer $l$. 
We note that these statistics are implicitly captured through the batch normalization in $f^{t-1}$ without storing input data for all previous tasks $\{1, ..., t-1\}$: $\mathbb{E}\big[\mu_l(x) | x \sim {\mathcal{X}_{\{1, ..., t-1\}}}\big] \simeq  \text{BN}_{l}(\text{running mean})$, $\mathbb{E}\big[\sigma_l(x) | x \sim {\mathcal{X}_{\{1, ..., t-1\}}}\big] \simeq  \text{BN}_{l}(\text{running var}).$
Otherwise, we approximate them using values calculated with post-convolution feature maps given a current task batch to the target model, $f^{t-1}$, leveraging its feature extraction capability for previous tasks. 
This efficient QDI allows us to use it for continual learning with minimal additional cost to existing methods. 
and the learning objective in \Eqref{eq:our_objective} can be updated as:
\begin{align}\label{eq:qdi_objective}
    {\underset{W_t}{\text{minimize}}}~~ \mathbb{E}_{(x_i, y_i)\sim \mathcal{D}_t}[\ell\left(f^t(x_i),  y^t_i(\psi)\right)] \nonumber \\ + \: \alpha \cdot \text{KL}\left(p^{t}_i(\tau), p^{t-1}_i(\tau)\right) + \: \beta \cdot \text{KL}\left(\tilde{p}^{t}_i(\tau), \tilde{p}^{t-1}_i(\tau)\right),
\end{align}
where $\tilde{p}^t(\tau)$ and $\tilde{p}^{t-1}(\tau)$ are the output probabilities of the generated examples scaled with temperature $\tau$ using the current and past task-models, $\beta$ is the hyper-parameter to control the strength of QDI knowledge distillation loss. QDI utilizes the generated examples to enhance knowledge distillation with a four times speedup to prior data-inversion methods~\citep{yin2020dreaming} due to the current task being a better prior than pixel-wise Gaussian to learn the generated data.

\section{Experiments}

\subsection{Experimental setup}

\noindent{\bf Datasets and architectures.} 
In this subsection, we provide the details about the datasets and the architectures, where we consider various architectural innovations in the ImageNet challenge~\citep{deng2009imagenet} for HCL.
\begin{enumerate}
    \item {\bf Split CIFAR-10~\citep{alex12cifar}.} It consists of CIFAR-10 sized $32 \times 32$ split into five tasks, where each task contains two different classes. We consider a different architecture for each task in HCL: LeNet~\citep{lecun89backpropagation}, ResNet~\citep{he2016deep}, DenseNet~\citep{huang2017densely}, SENet~\citep{hu2018squeeze}, and RegNet~\citep{xu2022regnet}. 
    \item {\bf Split CIFAR-100~\citep{alex12cifar}.} It is a split of CIFAR-100, where the 100 object classes are divided into 20 tasks. We consider nine architectures in the HCL: LeNet~\citep{lecun89backpropagation}, AlexNet~\citep{krizhevsky12nips}, VGG~\citep{simonyan2014very} with batch normalization~\citep{ioffe2015batch}, InceptionNet~\citep{szegedy2017inception}, ResNet~\citep{he2016deep}, ResNeXt~\citep{xie2017aggregated}, DenseNet~\citep{huang2017densely}, SENet~\citep{hu2018squeeze} for two tasks each and RegNet~\citep{xu2022regnet} for last four tasks.
    \item {\bf Split Tiny-ImageNet.} It is a variant of ImageNet~\citep{deng2009imagenet} that consists of images sized $64 \times 64$ from 200 classes. It consists of 10 tasks, where each task contains 20 classes. The set of architectures in HCL include LeNet~\citep{lecun89backpropagation}, ResNet~\citep{he2016deep}, ResNeXt~\citep{xie2017aggregated}, SENet~\citep{hu2018squeeze}, and RegNet~\citep{xu2022regnet} architectures, where each architecture is used for two consecutive tasks.
\end{enumerate}
\begin{table*}[t]

\setlength{\tabcolsep}{3pt} %
\resizebox{\textwidth}{!}{
\begin{tabular}{ll@{\hspace{6pt}}ccccccc}
\toprule
& {\textsc{Method}}& $\mathcal{B}$ & \multicolumn{2}{c}{\textsc{Split CIFAR-10}} &\multicolumn{2}{c}{\textsc{Split CIFAR-100}}&\multicolumn{2}{c}{\textsc{Split Tiny-ImageNet}}\\
\midrule
& & & $\mathcal{A}_T~(\uparrow)$ & $\mathcal{F}_T~(\downarrow)$ & $\mathcal{A}_T~(\uparrow)$ & $\mathcal{F}_T~(\downarrow)$ & $\mathcal{A}_T~(\uparrow)$ & $\mathcal{F}_T~(\downarrow)$\\
\midrule
& \multicolumn{8}{c}{\textsc{Standard Continual Learning}} \\
\midrule
& \textsc{Finetune} & -- & 
{62.93} \scriptsize($\pm$ 1.41) &  {41.45} \scriptsize($\pm$ 2.39) &
{40.13} \scriptsize($\pm$ 3.60) & {53.64} \scriptsize($\pm$ 3.65) &
{19.31} \scriptsize($\pm$ 0.71) &  {59.50} \scriptsize($\pm$ 0.81) \\

& \textsc{SI}~{\small \citep{zenke17si}}$^*$  & -- & 
{68.05} \scriptsize($\pm$ 5.91) &  {38.76} \scriptsize($\pm$ 0.89) &
{--} &  {--} &
{36.32} \scriptsize($\pm$ 0.13) &  {--} \\

& \textsc{LwF}~{\small \citep{LiZ2016learning}}$^*$  & -- &
{63.29} \scriptsize($\pm$ 2.35) &  {32.56} \scriptsize($\pm$ 0.56) &
{--} & {--} &
{15.85} \scriptsize($\pm$ 0.58) &  {--} \\

& \textsc{DI}~{\small \citep{yin2020dreaming}}  & -- & 
{89.11} \scriptsize($\pm$ 3.70) &  {7.12} \scriptsize($\pm$ 2.43) &
{49.64} \scriptsize($\pm$ 0.59) &  {35.29} \scriptsize($\pm$ 0.75) &
{44.29} \scriptsize($\pm$ 1.88) &  {22.42} \scriptsize($\pm$ 3.23) \\

\cmidrule{2-9}
\rowcolor{lgreen} & \textsc{KD (Ours)} & -- & 
{95.01} \scriptsize($\pm$ 0.79) & {0.87} \scriptsize($\pm$ 0.35) & %
{87.13} \scriptsize($\pm$ 0.45) & {4.26} \scriptsize($\pm$ 0.41) & %
{65.48} \scriptsize($\pm$ 0.12) & {7.33} \scriptsize($\pm$ 2.03) \\

\rowcolor{lgreen} & \textsc{KD w/ QDI (Ours)} & -- & 
{\bf 95.46} \scriptsize(\bf $\pm$ 0.15) & {\bf 0.37} \scriptsize(\bf $\pm$ 0.09) & %
{\bf 88.30} \scriptsize(\bf $\pm$ 0.17) & {\bf 2.00} \scriptsize($\pm$ \bf 0.15) & %
{\bf 66.79} \scriptsize($\pm$ \bf 0.45) & {\bf 7.28} \scriptsize($\pm$ \bf 0.49) \\

\cmidrule{2-9}

& \textsc{ICARL}~{\small \citep{rebuffi2017icarl}}$^*$  & \checkmark &
{88.99} \scriptsize($\pm$ 2.13) &  {2.63} \scriptsize($\pm$ 3.48) &
{--} & {--} &
{28.19} \scriptsize($\pm$ 1.47) &  {--} \\

& \textsc{A-GEM}~{\small \citep{chaudhry2018efficient}} & \checkmark & 
{87.09} \scriptsize($\pm$ 0.84) & {12.54} \scriptsize($\pm$ 1.23) &
{63.88} \scriptsize($\pm$ 1.23) & {29.98} \scriptsize($\pm$ 1.40) & 
{22.44} \scriptsize($\pm$ 0.24) &  {55.43} \scriptsize($\pm$ 0.71)\\

& \textsc{ER}~{\small ~{\small \citep{rolnick19er}}} & \checkmark & 
{91.39} \scriptsize($\pm$ 0.41) & {5.99} \scriptsize($\pm$ 0.43) &
{67.06} \scriptsize($\pm$ 1.01) & {25.31} \scriptsize($\pm$ 1.27) &
{21.03} \scriptsize($\pm$ 2.41) &  {57.88} \scriptsize($\pm$ 2.63) \\

& \textsc{DER}~{\small \citep{buzzega2020dark}} & \checkmark & 
92.45 \scriptsize($\pm$ 0.26) & {~~5.79} \scriptsize($\pm$ 0.20) & 
{67.74} \scriptsize($\pm$ 0.88) & {24.77} \scriptsize($\pm$ 0.94) & 
{30.97} \scriptsize($\pm$ 0.14) &  {45.25} \scriptsize($\pm$ 0.48) \\

& \textsc{DER++}~{\small \citep{buzzega2020dark}} &\checkmark & 
92.16 \scriptsize($\pm$ 0.68) & {~~5.96} \scriptsize($\pm$ 0.74) & 
{69.03} \scriptsize($\pm$ 0.74) & {23.06} \scriptsize($\pm$ 0.69) & 
{33.63} \scriptsize($\pm$ 0.10) &  {40.66} \scriptsize($\pm$ 0.83) \\

\cmidrule{2-9}

\rowcolor{lgreen} & \textsc{KD w/ Buffer (Ours)} & \checkmark & 
{\bf 95.81} \scriptsize(\bf $\pm$ 0.03) & {\bf 0.73} \scriptsize(\bf $\pm$ 0.34) & %
{\bf 80.35} \scriptsize(\bf $\pm$ 0.53) & {\bf 10.76} \scriptsize(\bf $\pm$ 0.65) & %
{\bf 69.07} \scriptsize(\bf $\pm$ 0.26) & {\bf 4.37} \scriptsize(\bf $\pm$ 0.17) \\

\cmidrule{2-9}
& \textsc{Multitask}$^*$ & \checkmark &  {98.31} \scriptsize($\pm$ 0.12) & \textsc{N/A} & {93.89} \scriptsize($\pm$ 0.78) & \textsc{N/A} & {82.04} \scriptsize($\pm$ 0.10) & \textsc{N/A} \\
\cmidrule{2-9}

& \multicolumn{8}{c}{\textsc{Heterogeneous Continual Learning}} \\
\midrule
& \textsc{Finetune} & -- & 
{61.99} \scriptsize($\pm$ 1.98) & {42.15} \scriptsize($\pm$ 2.94) & 
{27.08} \scriptsize($\pm$ 1.20) & {64.54} \scriptsize($\pm$ 1.45) & 
{14.83} \scriptsize($\pm$ 0.53) & {56.31} \scriptsize($\pm$ 0.60) \\

& \textsc{DI}~{\small \citep{yin2020dreaming}}  & -- & 
{81.16} \scriptsize($\pm$ 2.77) &  {15.84} \scriptsize($\pm$ 3.01) &
{38.33} \scriptsize($\pm$ 1.44) &  {47.08} \scriptsize($\pm$ 1.37) &
{31.69} \scriptsize($\pm$ 1.99) &  {32.42} \scriptsize($\pm$ 3.63) \\

\cmidrule{2-9}
\rowcolor{lgreen} & \textsc{KD (Ours)} & -- & 
{\bf 93.13} \scriptsize(\bf $\pm$ 0.35) & {\bf 3.97} \scriptsize(\bf $\pm$ 0.72) & %
{68.97} \scriptsize($\pm$ 0.72) & {21.45} \scriptsize($\pm$ 0.91) & %
{52.86} \scriptsize($\pm$ 2.78) & {11.73} \scriptsize($\pm$ 2.59) \\

\rowcolor{lgreen} & \textsc{KD w/ QDI (Ours)} & -- & 
{90.30} \scriptsize($\pm$ 1.27)  & {6.34} \scriptsize($\pm$ 0.15) &  

{\bf 76.44} \scriptsize(\bf $\pm$ 1.36) & {\bf 12.31} \scriptsize(\bf $\pm$ 0.99) & %
{\bf 53.39} \scriptsize(\bf $\pm$ 1.67) & {\bf 13.66} \scriptsize(\bf $\pm$ 1.18)\\
\cmidrule{2-9}

& \textsc{A-GEM}~{\small \citep{chaudhry2018efficient}} & \checkmark & 
{76.70} \scriptsize($\pm$ 0.67) & {24.31} \scriptsize($\pm$ 1.03) &
{35.37} \scriptsize($\pm$ 1.29) & {56.41} \scriptsize($\pm$ 2.19) & 
{15.79} \scriptsize($\pm$ 0.60) & {54.10} \scriptsize($\pm$ 1.00) \\

& \textsc{ER}~{\small ~{\small \citep{rolnick19er}}} &\checkmark & 
{81.78} \scriptsize($\pm$ 2.15) & {17.57} \scriptsize($\pm$ 2.86) &
{52.59} \scriptsize($\pm$ 1.03) & {36.41} \scriptsize($\pm$ 0.54) & 
{27.82} \scriptsize($\pm$ 1.43) & {39.66} \scriptsize($\pm$ 1.24) \\

& \textsc{DER}~{\small \citep{buzzega2020dark}} & \checkmark & 
{79.92} \scriptsize($\pm$ 0.51) & {19.99} \scriptsize($\pm$ 0.64) & %
{53.01} \scriptsize($\pm$ 0.61) & {37.33} \scriptsize($\pm$ 0.58) & %
{25.27} \scriptsize($\pm$ 1.34) & {40.81} \scriptsize($\pm$ 0.57) \\

& \textsc{DER++}~{\small \citep{buzzega2020dark}} & \checkmark & 
{81.33} \scriptsize($\pm$ 0.76) & {18.79} \scriptsize($\pm$ 0.75) & %
{54.81} \scriptsize($\pm$ 1.65) & {35.20} \scriptsize($\pm$ 1.63) & %
{29.74} \scriptsize($\pm$ 1.40) & {36.79} \scriptsize($\pm$ 2.16)\\

\cmidrule{2-9}

\rowcolor{lgreen} & \textsc{KD w/ Buffer (Ours)} & \checkmark & 
{\bf 93.95} \scriptsize(\bf $\pm$ 0.22) & {\bf 2.49} \scriptsize(\bf $\pm$ 0.44) & %
{\bf 73.98} \scriptsize(\bf $\pm$ 0.13) & {\bf 16.40} \scriptsize(\bf $\pm$ 0.18) & %
{\bf 54.84} \scriptsize(\bf $\pm$ 0.87) & {\bf 10.82} \scriptsize(\bf $\pm$ 0.80) \\

\bottomrule
    \end{tabular}}
    \vspace{-0.1in}
    \caption{{\bf Accuracy and forgetting} with task-IL on standard CL and HCL. The best results are highlighted in {\bf bold}. $\mathcal{B}$ denotes replay-buffer, $\mathcal{A}_T, \mathcal{F}_T$ denote average accuracy and forgetting after the completion of training.\textcolor{black}{$^*$~denotes the methods whose numbers were used from \citet{buzzega2020dark} and $-$ indicates the unavailability of results. All other experiments are over three independent runs.}\label{tab:main_tiltable}}
\end{table*}

\noindent{\bf Baselines.} We compare our designed \textsc{KD} paradigm (\Eqref{eq:our_objective}), with \textsc{QDI} (\Eqref{eq:qdi_objective}) agains standard fine-tuning ($\textsc{Finetune}$), \textsc{SI}~\cite{zenke17si}, \textsc{LwF}~\citep{LiZ2016learning}, \textsc{ICARL}~\citep{rebuffi2017icarl} \textsc{A-GEM}~\citep{chaudhry2018efficient}, \textsc{ER}~\citep{chaudhry2019continual}, \textsc{DER} and \textsc{DER}++~\citep{buzzega2020dark}, \textsc{DI}~\citep{yin2020dreaming} and multi task training ($\textsc{Multitask}$), where all tasks are trained jointly with ResNet18 -- most prominent architecture in CL literature. For a fair comparison, we extend our method with the classification loss using \textsc{Buffer} in \Eqref{eq:our_objective}.
The buffer-size for replay-based method is fixed to $200$ for all the methods that use exemplar or generative replay. We use the same set of augmentations for training all the methods.

\paragraph{Evaluation.} To evaluate and compare our method in both standard CL and HCL setting, we use the average accuracy and forgetting as two metrics.
\begin{enumerate}
\item {\bf Average accuracy.} It is the average of the accuracy of all tasks after the completion of training with the sequence of $T$ tasks. More formally, average accuracy $\mathcal{A}_T = \frac{1}{T}\sum_{t=1}^T a_{T, t}$, where $a_{i, j}$ is the accuracy of task $j$ after completion of task $i$.

\item {\bf Average forgetting.} It is the averaged difference in accuracy between the accuracy after training and the maximum accuracy for each task. More formally, average forgetting $\mathcal{F}_T~=~\frac{1}{T-1}\sum_{t=1}^T\underset{i \in \{1, \ldots, T\}}{\max}\left(a_{i, t} - a_{T, t}\right)$
\end{enumerate}

\begin{table*}[t!]
\begin{minipage}[t]{0.48\linewidth}
    \centering
    \resizebox{.9\linewidth}{!}{%
    \begin{tabular}{lcccc} 
    \toprule
     \textsc{Model}  & \textsc{Task-IL} & \textsc{Class-IL} \\ \midrule
      \textsc{KD}  & {83.43} \scriptsize($\pm$ 0.42) &   {21.11} \scriptsize($\pm$ 0.19) \\
        $+$ \textcolor{color3}{\textsc{AUG-KD}}  & {91.76} \scriptsize($\pm$ 2.03) &   {34.11} \scriptsize($\pm$ 1.07) \\
        ~$+$ \textcolor{color3}{\textsc{LS}} & {93.13} \scriptsize($\pm$ 0.35) &   {30.21} \scriptsize($\pm$ 0.11)\\
        \midrule
        ~~$+$\textcolor{color4}{\textsc{Warmup}}  & {91.64} \scriptsize($\pm$ 1.37) &   {28.26} \scriptsize($\pm$ 0.28) \\
        ~~~$+$ \textcolor{color4}{\textsc{TI}}  & {90.14} \scriptsize($\pm$ 1.33) &   {22.04} \scriptsize($\pm$ 0.27)  \\
 \bottomrule
   \end{tabular}}
    \vspace{-0.1in}
            \captionof{table}{ {\bf Ablation} showin $\mathcal{A}_t$ to measure the effect of various components with Split-CIFAR10. The inclusion of distillation with augmented images (\textsc{aug-kd}) and label-smoothing (\textsc{ls)} significantly improve performance, whereas \textsc{warmup} and task-identity~(\textsc{ti)} degrade performance. \textcolor{color3}{Blue} and \textcolor{color4}{red} colors denote the components that were included and excluded from the method respectively.\label{tab:l2_u}}
        \centering
    \resizebox{.95\linewidth}{!}{%
    \begin{tabular}{lcccc} 
    \toprule
     \textsc{Model}  & \textsc{Task-IL} & \textsc{Class-IL} \\ \midrule
        \textsc{CE} & {91.06} \scriptsize($\pm$ 0.63) & {20.99} \scriptsize($\pm$ 0.27)  & \\
      \textsc{MSE} & {89.68} \scriptsize($\pm$ 0.99) & {20.09} \scriptsize($\pm$ 0.26) \\
      \textsc{KL}  & 93.13 \scriptsize($\pm$ 0.35) &   {30.21} \scriptsize($\pm$ 0.11) \\
 \bottomrule
    \end{tabular}}
    \vspace{-0.1in}
            \captionof{table}{{\bf Ablation} showing $\mathcal{A}_t$ using validation set to measure the effect of different distance metrics. We measure the average accuracy and forgetting for HCL on Split-CIFAR10 dataset. \label{tab:distance_table}}
\end{minipage}\hfill
\begin{minipage}[t]{0.48\linewidth}
    \centering
    \resizebox{.85\linewidth}{!}{%
    \small
    \begin{tabular}{lcccc} 
    \toprule
     \textsc{Model}  & $\mathcal{A}_T~(\uparrow)$ & $\mathcal{F}_T~(\downarrow)$ \\ \midrule
        \textsc{Finetune} & {55.47} \scriptsize($\pm$ 2.06) & {77.68} \scriptsize($\pm$ 3.68)  & \\
      \textsc{Der}++ & {69.05} \scriptsize($\pm$ 2.56) & {51.99} \scriptsize($\pm$ 1.91)  & \\
      \midrule
      \textsc{KD (Ours)}  & {88.77} \scriptsize($\pm$ 0.55) &   {10.95} \scriptsize($\pm$ 1.34) \\
      \midrule
    \textsc{Multitask} & {88.71} \scriptsize($\pm$ 0.40) & {--} \\
 \bottomrule
 \vspace{1mm}
    \end{tabular}}
    \vspace{-0.3in}
    \centering
            \captionof{table}{{\bf Transformer analysis.}  Average accuracy and forgetting for HCL with ResNet18 to Swin-T with two tasks on CIFAR-10. \label{tab:swin_table}}
           \centering
            \includegraphics[width=0.95\linewidth]{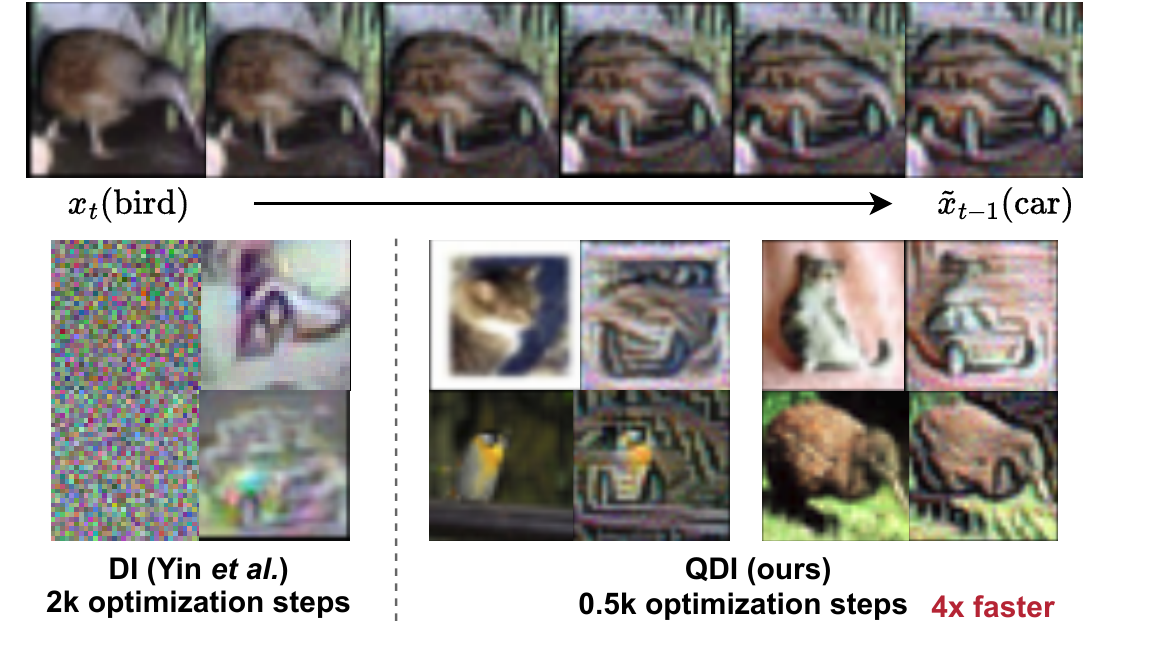}
    \vspace{-0.15in}
           \captionof{figure}{ Visualization of QDI and comparison to prior art. {\bf Top:} prior feature synthesis from current task data. {\bf Bottom:} samples for DI that start from $\mathcal{N}(0, 1)$ and QDI from $x_{t}$ in pairs (left -- optimization start point; right -- optimization convergence point) for ResNet18 on CIFAR-10 dataset. 
           \label{fig:di_qdi} }
\end{minipage}

\vspace{-0.15in}
\end{table*}
\subsection{Quantitative results}

\noindent{\bf Task-incremental continual learning.} First, we compare the performance of our proposed method with other baselines for task-IL on standard and heterogeneous continual learning settings. In \Cref{tab:main_tiltable}, we find that the performance of all the methods drop significantly by going from the SCL to the HCL setting. Contrary, our proposed methods achieve the lowest drop across all methods; moreover, it obtains the best performance with and without buffer across all considered scenarios. Additionally, we show that in standard CL and HCL, QDI leads to a $\sim40\%$ and $\sim 22\%$ absolute improvement in average accuracy over state-of-the-art data-free methods for Split CIFAR-100 and Tiny-ImageNet. Interestingly, we find that distillation with synthetic examples generated by QDI obtains comparable or better performance to the replay buffer highlighting the flexibility of data-free continual learning.

Moreover, QDI improves synthesis speed by $4\times$ for generating the synthetic examples in comparison to DI~\citep{yin2020dreaming} for each task. 
All variants of our method achieve the least forgetting in comparison to the baselines across all datasets for both the standard CL and HCL scenarios. Our proposed training scheme is especially beneficial when past-task data is unavailable due to privacy concerns. It allows the adaptation to the new network structure for the new task by utilizing the saved model checkpoints. It enables training with different network structures without storing any past-task instances.

\paragraph{Class-incremental continual learning.}
We also show the validation of our proposed method on class-IL in the supplementary material, where task-labels are not provided during training or inference. We observe that all our methods outperform the baselines with and without buffer in average accuracy and forgetting. Specifically, we show that ours with \textsc{QDI} and \textsc{Buffer} examples show a five to ten percent improvement in average accuracy for Split CIFAR-100 and Tiny-ImageNet datasets. However, we find that buffer is an important component of our method for class-IL; it is  consistent to the observations in prior works~\citep{yin2020dreaming, carta2022ex}. Further, we observe that the synthetic examples generated by QDI lead to unstable training for CIFAR-10, but improve the performance for other datasets. We believe this is an outcome of the model's dependence on domain over semantics~\citep{smith2021always} and further investigation of data-free continual learning techniques for class-IL would be an
interesting direction for future work.

\subsection{Additional analysis}
\noindent{\bf Ablation studies.} To dissect the contribution of different components in our proposed method for HCL, we isolate and study them on Split CIFAR-10 in \Cref{tab:l2_u}. First, we find that knowledge distillation with augmented instances, in the contrast to a common setting of precomputed logits,
improves the performance by $\sim8\%$ and $\sim13\%$ for task-IL and class-IL respectively due to the better generalization of current task representations. Next, we observe that label-smoothing improves the task-IL performance while hurting the class-IL average accuracy highlighting the presence of a trade-off between both the settings. Additionally, we measure the effect of warmup and task-identity for HCL following prior works~\citep{jung2016less, LiZ2016learning} for standard CL, where we find that warm-up of the linear head before current-task training and inclusion of task-identity hurt the performance for HCL. 

Further, we conduct an ablation on different distance metrics used for KD in \Eqref{eq:our_objective} for HCL. While, a modified version of cross entropy (CE) or mean-square distance (MSE) are commonly used for knowledge distillation in CL~\citep{LiZ2016learning, rebuffi2017icarl, buzzega2020dark}, we find the KL-divergence (KL) is more stable and often obtains better performance in \Cref{tab:distance_table} for both class-IL and task-IL. 
We believe that the ablation of these design choices will provide critical insights towards building future methods for standard and heterogeneous CL.

\noindent{\bf Transfer to vision transformer.} To show the flexibility of our framework beyond convolutional architectures, we design a two task experiment with CIFAR-10 consisting of five classes each. The first task is trained with ResNet18~\citep{he2016deep} followed by Swin-T~\citep{liu2021swin} on the second task. \Cref{tab:swin_table} shows the comparison of our proposed method with only KD against \textsc{Finetune}, \textsc{DER++} trained with HCL and \textsc{Multitask} learned with CIFAR-10 on Swin-T~\citep{liu2021swin}. First, we remark that Swin-T (\textsc{Multitask}: $88.71 \pm 0.40$) achieves lower accuracy in comparison to ResNet18 (\textsc{Multitask}: $98.31 \pm 0.12$) because ViTs are data-hungry~\citep{liu2021efficient}. Second, we show that our proposed method with evolving architectures obtains performance similar to the multitask baseline. Lastly, our proposed method obtains better average accuracy and lower forgetting over the evaluated baselines showing the benefit of training with multiple architectures. 

\noindent{\bf Visualization of generated examples.}
We visualize the generated examples by DI and QDI in the HCL setting for ResNet18 architecture with CIFAR-10 dataset in \Cref{fig:di_qdi}. To explain the interplay of the generated examples initialized with the current task instances and the past task examples, we consider the example of a current task (\textit{e.g.,} a cat or bird) to invert the visual features of the past task (car). We observe that the input-features swiftly generated by QDI have a combination of both classes and superior in realism; in contrast, DI only contains the past-task features and requires more optimization steps from random noise for full convergence. We believe this interpolation also helps in current task-training while also alleviating catastrophic forgetting.

\section{Conclusion}
In this work, we propose a novel continual learning setting named \emph{Heterogeneous Continual Learning (HCL)}, where the lifelong learner learns on a sequence of tasks while adopting state-of-the-art deep-learning techniques and architectures. We benchmark the state-of-the-art CL solutions in this new setting and observe a large degradation in performance. To tackle this limitation, we revisit knowledge distillation and propose a modified paradigm inspired by the recent advances in knowledge distillation. Additionally, we propose \emph{Quick Deep Inversion (QDI)} that generates synthetic examples using current task instances to enhance knowledge distillation performance in the data-free continual learning setup at a negligible additional cost. Our experimental evaluation shows that without tuning the training configurations, we achieve a significant improvement for both the standard and heterogeneous continual learning scenarios across all datasets, highlighting our proposed approach's efficacy in real-world applications.

\noindent{\bf Limitations.} \label{sec:limitations}
Although the proposed approach is applicable for general deep-learning architectures, we found some limitations with the current approach/setup. First, in some cases, the recent architectures can be sub-optimal and degrade the model performance due to their training configuration, model size, or hyper-parameter mismatch. 
We note that in this work, we did not tune the training configuration or hyper-parameters for each model for fair comparisons. However, model specific hyper-parameter tuning can further improve model performance. 
Second, our work is not yet applicable to unsupervised CL with heterogeneous architectures, where the learner is expected to learn on unlabelled data with changing architectures because model inversion and KD with unsupervised CL is not a trivial extension due to the absence of class-dependent information to generate class-conditioned examples.

\bibliography{references}

\begin{thebibliography}{59}
\providecommand{\natexlab}[1]{#1}
\providecommand{\url}[1]{\texttt{#1}}
\expandafter\ifx\csname urlstyle\endcsname\relax
  \providecommand{\doi}[1]{doi: #1}\else
  \providecommand{\doi}{doi: \begingroup \urlstyle{rm}\Url}\fi

\bibitem[Ahn et~al.(2019)Ahn, Cha, Lee, and Moon]{ahn2019uncertainty}
Hongjoon Ahn, Sungmin Cha, Donggyu Lee, and Taesup Moon.
\newblock Uncertainty-based continual learning with adaptive regularization.
\newblock In \emph{Advances in Neural Information Processing Systems
  (NeurIPS)}, 2019.

\bibitem[Aljundi et~al.(2019)Aljundi, Lin, Goujaud, and
  Bengio]{Aljundi2019GradientBS}
Rahaf Aljundi, Min Lin, Baptiste Goujaud, and Yoshua Bengio.
\newblock Gradient based sample selection for online continual learning.
\newblock In \emph{Advances in Neural Information Processing Systems
  (NeurIPS)}, 2019.

\bibitem[Beyer et~al.(2022)Beyer, Zhai, Royer, Markeeva, Anil, and
  Kolesnikov]{beyer2022knowledge}
Lucas Beyer, Xiaohua Zhai, Am{\'e}lie Royer, Larisa Markeeva, Rohan Anil, and
  Alexander Kolesnikov.
\newblock Knowledge distillation: A good teacher is patient and consistent.
\newblock In \emph{Proceedings of the IEEE International Conference on Computer
  Vision and Pattern Recognition (CVPR)}, 2022.

\bibitem[Buzzega et~al.(2020)Buzzega, Boschini, Porrello, Abati, and
  Calderara]{buzzega2020dark}
Pietro Buzzega, Matteo Boschini, Angelo Porrello, Davide Abati, and Simone
  Calderara.
\newblock Dark experience for general continual learning: a strong, simple
  baseline.
\newblock In \emph{Advances in Neural Information Processing Systems
  (NeurIPS)}, 2020.

\bibitem[Carta et~al.(2022)Carta, Cossu, Lomonaco, and Bacciu]{carta2022ex}
Antonio Carta, Andrea Cossu, Vincenzo Lomonaco, and Davide Bacciu.
\newblock Ex-model: Continual learning from a stream of trained models.
\newblock In \emph{Proceedings of the IEEE International Conference on Computer
  Vision and Pattern Recognition (CVPR) Workshops}, 2022.

\bibitem[Chandrasegaran et~al.(2022)Chandrasegaran, Tran, Zhao, and
  Cheung]{chandrasegaran2022revisiting}
Keshigeyan Chandrasegaran, Ngoc-Trung Tran, Yunqing Zhao, and Ngai-Man Cheung.
\newblock Revisiting label smoothing and knowledge distillation compatibility:
  What was missing?
\newblock In \emph{Proceedings of the International Conference on Machine
  Learning (ICML)}, 2022.

\bibitem[Chaudhry et~al.(2019{\natexlab{a}})Chaudhry, Ranzato, Rohrbach, and
  Elhoseiny]{chaudhry2018efficient}
Arslan Chaudhry, Marc'Aurelio Ranzato, Marcus Rohrbach, and Mohamed Elhoseiny.
\newblock Efficient lifelong learning with a-gem.
\newblock In \emph{Proceedings of the International Conference on Learning
  Representations (ICLR)}, 2019{\natexlab{a}}.

\bibitem[Chaudhry et~al.(2019{\natexlab{b}})Chaudhry, Rohrbach, Elhoseiny,
  Ajanthan, Dokania, Torr, and Ranzato]{chaudhry2019continual}
Arslan Chaudhry, Marcus Rohrbach, Mohamed Elhoseiny, Thalaiyasingam Ajanthan,
  Puneet~K Dokania, Philip~HS Torr, and M~Ranzato.
\newblock Continual learning with tiny episodic memories.
\newblock \emph{arXiv preprint arXiv:1902.10486}, 2019{\natexlab{b}}.

\bibitem[Cossu et~al.(2022)Cossu, Graffieti, Pellegrini, Maltoni, Bacciu,
  Carta, and Lomonaco]{cossu2022class}
Andrea Cossu, Gabriele Graffieti, Lorenzo Pellegrini, Davide Maltoni, Davide
  Bacciu, Antonio Carta, and Vincenzo Lomonaco.
\newblock Is class-incremental enough for continual learning?
\newblock \emph{Frontiers in Artificial Intelligence}, 2022.

\bibitem[Dai et~al.(2020)Dai, Yin, and Jha]{dai2020incremental}
Xiaoliang Dai, Hongxu Yin, and Niraj~K Jha.
\newblock Incremental learning using a grow-and-prune paradigm with efficient
  neural networks.
\newblock \emph{IEEE Transactions on Emerging Topics in Computing}, 2020.

\bibitem[De~Lange et~al.(2021)De~Lange, Aljundi, Masana, Parisot, Jia,
  Leonardis, Slabaugh, and Tuytelaars]{de2021continual}
Matthias De~Lange, Rahaf Aljundi, Marc Masana, Sarah Parisot, Xu~Jia,
  Ale{\v{s}} Leonardis, Gregory Slabaugh, and Tinne Tuytelaars.
\newblock A continual learning survey: Defying forgetting in classification
  tasks.
\newblock \emph{IEEE Transactions on Pattern Analysis and Machine
  Intelligence}, 2021.

\bibitem[Deng et~al.(2009)Deng, Dong, Socher, Li, Li, and
  Fei-Fei]{deng2009imagenet}
Jia Deng, Wei Dong, Richard Socher, Li-Jia Li, Kai Li, and Li~Fei-Fei.
\newblock Imagenet: A large-scale hierarchical image database.
\newblock In \emph{Proceedings of the IEEE International Conference on Computer
  Vision and Pattern Recognition (CVPR)}, 2009.

\bibitem[Dhar et~al.(2019)Dhar, Singh, Peng, Wu, and
  Chellappa]{dhar2019learning}
Prithviraj Dhar, Rajat~Vikram Singh, Kuan-Chuan Peng, Ziyan Wu, and Rama
  Chellappa.
\newblock Learning without memorizing.
\newblock In \emph{Proceedings of the IEEE International Conference on Computer
  Vision and Pattern Recognition (CVPR)}, 2019.

\bibitem[Dosovitskiy et~al.(2021)Dosovitskiy, Beyer, Kolesnikov, Weissenborn,
  Zhai, Unterthiner, Dehghani, Minderer, Heigold, Gelly, Uszkoreit, and
  Houlsby]{dosovitskiy2021an}
Alexey Dosovitskiy, Lucas Beyer, Alexander Kolesnikov, Dirk Weissenborn,
  Xiaohua Zhai, Thomas Unterthiner, Mostafa Dehghani, Matthias Minderer, Georg
  Heigold, Sylvain Gelly, Jakob Uszkoreit, and Neil Houlsby.
\newblock An image is worth 16x16 words: Transformers for image recognition at
  scale.
\newblock In \emph{Proceedings of the International Conference on Learning
  Representations (ICLR)}, 2021.

\bibitem[He et~al.(2016)He, Zhang, Ren, and Sun]{he2016deep}
Kaiming He, Xiangyu Zhang, Shaoqing Ren, and Jian Sun.
\newblock Deep residual learning for image recognition.
\newblock In \emph{Proceedings of the IEEE International Conference on Computer
  Vision and Pattern Recognition (CVPR)}, 2016.

\bibitem[Hinton et~al.(2015)Hinton, Vinyals, and Dean]{hinton2015distilling}
Geoffrey Hinton, Oriol Vinyals, and Jeff Dean.
\newblock Distilling the knowledge in a neural network.
\newblock \emph{arXiv preprint arXiv:1503.02531}, 2015.

\bibitem[Hou et~al.(2019)Hou, Pan, Loy, Wang, and Lin]{Hou19learning}
Saihui Hou, Xinyu Pan, Chen~Change Loy, Zilei Wang, and Dahua Lin.
\newblock Learning a unified classifier incrementally via rebalancing.
\newblock In \emph{Proceedings of the IEEE International Conference on Computer
  Vision and Pattern Recognition (CVPR)}, 2019.

\bibitem[Howard et~al.(2017)Howard, Zhu, Chen, Kalenichenko, Wang, Weyand,
  Andreetto, and Adam]{howard2017mobilenets}
Andrew~G Howard, Menglong Zhu, Bo~Chen, Dmitry Kalenichenko, Weijun Wang,
  Tobias Weyand, Marco Andreetto, and Hartwig Adam.
\newblock Mobilenets: Efficient convolutional neural networks for mobile vision
  applications.
\newblock \emph{arXiv preprint arXiv:1704.04861}, 2017.

\bibitem[Hu et~al.(2018)Hu, Shen, and Sun]{hu2018squeeze}
Jie Hu, Li~Shen, and Gang Sun.
\newblock Squeeze-and-excitation networks.
\newblock In \emph{Proceedings of the IEEE International Conference on Computer
  Vision and Pattern Recognition (CVPR)}, 2018.

\bibitem[Huang et~al.(2017)Huang, Liu, Van Der~Maaten, and
  Weinberger]{huang2017densely}
Gao Huang, Zhuang Liu, Laurens Van Der~Maaten, and Kilian~Q Weinberger.
\newblock Densely connected convolutional networks.
\newblock In \emph{Proceedings of the IEEE International Conference on Computer
  Vision and Pattern Recognition (CVPR)}, 2017.

\bibitem[Ioffe \& Szegedy(2015)Ioffe and Szegedy]{ioffe2015batch}
Sergey Ioffe and Christian Szegedy.
\newblock Batch normalization: Accelerating deep network training by reducing
  internal covariate shift.
\newblock In \emph{Proceedings of the International Conference on Machine
  Learning (ICML)}, 2015.

\bibitem[Jung et~al.(2016)Jung, Ju, Jung, and Kim]{jung2016less}
Heechul Jung, Jeongwoo Ju, Minju Jung, and Junmo Kim.
\newblock Less-forgetting learning in deep neural networks.
\newblock \emph{arXiv preprint arXiv:1607.00122}, 2016.

\bibitem[Krizhevsky(2012)]{alex12cifar}
Alex Krizhevsky.
\newblock Learning multiple layers of features from tiny images.
\newblock \emph{University of Toronto}, 05 2012.

\bibitem[Krizhevsky et~al.(2012)Krizhevsky, Sutskever, and
  Hinton]{krizhevsky12nips}
Alex Krizhevsky, Ilya Sutskever, and Geoffrey~E Hinton.
\newblock Imagenet classification with deep convolutional neural networks.
\newblock In \emph{Advances in Neural Information Processing Systems
  (NeurIPS)}, 2012.

\bibitem[LeCun et~al.(1989)LeCun, Boser, Denker, Henderson, Howard, Hubbard,
  and Jackel]{lecun89backpropagation}
Y.~LeCun, B.~Boser, J.~S. Denker, D.~Henderson, R.~E. Howard, W.~Hubbard, and
  L.~D. Jackel.
\newblock Backpropagation applied to handwritten zip code recognition.
\newblock \emph{Neural Computation}, 1989.

\bibitem[LeCun(1998)]{lecun1998mnist}
Yann LeCun.
\newblock The mnist database of handwritten digits.
\newblock \emph{http://yann. lecun. com/exdb/mnist/}, 1998.

\bibitem[LeCun et~al.(1998)LeCun, Bottou, Bengio, and
  Haffner]{lecun1998gradient}
Yann LeCun, L{\'e}on Bottou, Yoshua Bengio, and Patrick Haffner.
\newblock Gradient-based learning applied to document recognition.
\newblock \emph{Proceedings of the IEEE}, 1998.

\bibitem[Lee et~al.(2021)Lee, Behpour, and Eaton]{lee21sharing}
Seungwon Lee, Sima Behpour, and Eric Eaton.
\newblock Sharing less is more: Lifelong learning in deep networks with
  selective layer transfer.
\newblock In \emph{Proceedings of the 38th International Conference on Machine
  Learning}, 2021.

\bibitem[Li et~al.(2019)Li, Zhou, Wu, Socher, and Xiong]{li2019learn}
Xilai Li, Yingbo Zhou, Tianfu Wu, Richard Socher, and Caiming Xiong.
\newblock Learn to grow: A continual structure learning framework for
  overcoming catastrophic forgetting.
\newblock In \emph{Proceedings of the International Conference on Machine
  Learning (ICML)}, 2019.

\bibitem[Li \& Hoiem(2016)Li and Hoiem]{LiZ2016learning}
Zhizhong Li and Derek Hoiem.
\newblock Learning without forgetting.
\newblock In \emph{Proceedings of the European Conference on Computer Vision
  (ECCV)}, 2016.

\bibitem[Liu et~al.(2021{\natexlab{a}})Liu, Sangineto, Bi, Sebe, Lepri, and
  De~Nadai]{liu2021efficient}
Yahui Liu, Enver Sangineto, Wei Bi, Nicu Sebe, Bruno Lepri, and Marco De~Nadai.
\newblock Efficient training of visual transformers with small-size datasets.
\newblock \emph{arXiv preprint arXiv:2106.03746}, 2021{\natexlab{a}}.

\bibitem[Liu et~al.(2021{\natexlab{b}})Liu, Lin, Cao, Hu, Wei, Zhang, Lin, and
  Guo]{liu2021swin}
Ze~Liu, Yutong Lin, Yue Cao, Han Hu, Yixuan Wei, Zheng Zhang, Stephen Lin, and
  Baining Guo.
\newblock Swin transformer: Hierarchical vision transformer using shifted
  windows.
\newblock In \emph{Proceedings of the International Conference on Computer
  Vision (ICCV)}, 2021{\natexlab{b}}.

\bibitem[Madaan et~al.(2022)Madaan, Yoon, Li, Liu, and
  Hwang]{madaan2022representational}
Divyam Madaan, Jaehong Yoon, Yuanchun Li, Yunxin Liu, and Sung~Ju Hwang.
\newblock Representational continuity for unsupervised continual learning.
\newblock In \emph{Proceedings of the International Conference on Learning
  Representations (ICLR)}, 2022.

\bibitem[McGrath et~al.(2021)McGrath, Kapishnikov, Toma{\v{s}}ev, Pearce,
  Hassabis, Kim, Paquet, and Kramnik]{mcgrath2021acquisition}
Thomas McGrath, Andrei Kapishnikov, Nenad Toma{\v{s}}ev, Adam Pearce, Demis
  Hassabis, Been Kim, Ulrich Paquet, and Vladimir Kramnik.
\newblock Acquisition of chess knowledge in alphazero.
\newblock \emph{arXiv preprint arXiv:2111.09259}, 2021.

\bibitem[Mehta et~al.(2021)Mehta, Ghazvininejad, Iyer, Zettlemoyer, and
  Hajishirzi]{mehta2021delight}
Sachin Mehta, Marjan Ghazvininejad, Srinivasan Iyer, Luke Zettlemoyer, and
  Hannaneh Hajishirzi.
\newblock Delight: Deep and light-weight transformer.
\newblock In \emph{Proceedings of the International Conference on Learning
  Representations (ICLR)}, 2021.

\bibitem[Mirzadeh et~al.(2022{\natexlab{a}})Mirzadeh, Chaudhry, Hu, Pascanu,
  Gorur, and Farajtabar]{mirzadeh2021wide}
Seyed~Iman Mirzadeh, Arslan Chaudhry, Huiyi Hu, Razvan Pascanu, Dilan Gorur,
  and Mehrdad Farajtabar.
\newblock Wide neural networks forget less catastrophically.
\newblock In \emph{Proceedings of the International Conference on Machine
  Learning (ICML)}, 2022{\natexlab{a}}.

\bibitem[Mirzadeh et~al.(2022{\natexlab{b}})Mirzadeh, Chaudhry, Yin, Hu,
  Pascanu, Gorur, and Farajtabar]{mirzadeh2022wide}
Seyed~Iman Mirzadeh, Arslan Chaudhry, Dong Yin, Huiyi Hu, Razvan Pascanu, Dilan
  Gorur, and Mehrdad Farajtabar.
\newblock Wide neural networks forget less catastrophically.
\newblock In \emph{Proceedings of the International Conference on Machine
  Learning (ICML)}, 2022{\natexlab{b}}.

\bibitem[Mirzadeh et~al.(2022{\natexlab{c}})Mirzadeh, Chaudhry, Yin, Nguyen,
  Pascanu, Gorur, and Farajtabar]{mirzadeh2022architecture}
Seyed~Iman Mirzadeh, Arslan Chaudhry, Dong Yin, Timothy Nguyen, Razvan Pascanu,
  Dilan Gorur, and Mehrdad Farajtabar.
\newblock Architecture matters in continual learning.
\newblock \emph{arXiv preprint arXiv:2202.00275}, 2022{\natexlab{c}}.

\bibitem[M{\"u}ller et~al.(2019)M{\"u}ller, Kornblith, and
  Hinton]{muller2019does}
Rafael M{\"u}ller, Simon Kornblith, and Geoffrey~E Hinton.
\newblock When does label smoothing help?
\newblock \emph{Neurips}, 2019.

\bibitem[Ramesh et~al.(2021)Ramesh, Pavlov, Goh, Gray, Voss, Radford, Chen, and
  Sutskever]{ramesh2021zero}
Aditya Ramesh, Mikhail Pavlov, Gabriel Goh, Scott Gray, Chelsea Voss, Alec
  Radford, Mark Chen, and Ilya Sutskever.
\newblock Zero-shot text-to-image generation.
\newblock In \emph{Proceedings of the International Conference on Machine
  Learning (ICML)}, 2021.

\bibitem[Rebuffi et~al.(2017)Rebuffi, Kolesnikov, Sperl, and
  Lampert]{rebuffi2017icarl}
Sylvestre-Alvise Rebuffi, Alexander Kolesnikov, Georg Sperl, and Christoph~H
  Lampert.
\newblock icarl: Incremental classifier and representation learning.
\newblock In \emph{Proceedings of the IEEE International Conference on Computer
  Vision and Pattern Recognition (CVPR)}, 2017.

\bibitem[Rolnick et~al.(2019)Rolnick, Ahuja, Schwarz, Lillicrap, and
  Wayne]{rolnick19er}
David Rolnick, Arun Ahuja, Jonathan Schwarz, Timothy Lillicrap, and Gregory
  Wayne.
\newblock Experience replay for continual learning.
\newblock In \emph{Advances in Neural Information Processing Systems
  (NeurIPS)}, 2019.

\bibitem[Rusu et~al.(2016)Rusu, Rabinowitz, Desjardins, Soyer, Kirkpatrick,
  Kavukcuoglu, Pascanu, and Hadsell]{rusu2016progressive}
Andrei~A Rusu, Neil~C Rabinowitz, Guillaume Desjardins, Hubert Soyer, James
  Kirkpatrick, Koray Kavukcuoglu, Razvan Pascanu, and Raia Hadsell.
\newblock Progressive neural networks.
\newblock \emph{arXiv preprint arXiv:1606.04671}, 2016.

\bibitem[Schwarz et~al.(2018)Schwarz, Luketina, Czarnecki, Grabska-Barwinska,
  Teh, Pascanu, and Hadsell]{schwarz2018progress}
Jonathan Schwarz, Jelena Luketina, Wojciech~M Czarnecki, Agnieszka
  Grabska-Barwinska, Yee~Whye Teh, Razvan Pascanu, and Raia Hadsell.
\newblock Progress \& compress: A scalable framework for continual learning.
\newblock In \emph{Proceedings of the International Conference on Machine
  Learning (ICML)}, 2018.

\bibitem[Shen et~al.(2021)Shen, Liu, Xu, Chen, Cheng, and Savvides]{shen2021is}
Zhiqiang Shen, Zechun Liu, Dejia Xu, Zitian Chen, Kwang-Ting Cheng, and Marios
  Savvides.
\newblock Is label smoothing truly incompatible with knowledge distillation: An
  empirical study.
\newblock In \emph{iclr}, 2021.

\bibitem[Simonyan \& Zisserman(2015)Simonyan and Zisserman]{simonyan2014very}
Karen Simonyan and Andrew Zisserman.
\newblock Very deep convolutional networks for large-scale image recognition.
\newblock In \emph{Proceedings of the International Conference on Learning
  Representations (ICLR)}, 2015.

\bibitem[Smith et~al.(2021)Smith, Hsu, Balloch, Shen, Jin, and
  Kira]{smith2021always}
James Smith, Yen-Chang Hsu, Jonathan Balloch, Yilin Shen, Hongxia Jin, and
  Zsolt Kira.
\newblock Always be dreaming: A new approach for data-free class-incremental
  learning.
\newblock In \emph{Proceedings of the International Conference on Computer
  Vision (ICCV)}, 2021.

\bibitem[Szegedy et~al.(2017)Szegedy, Ioffe, Vanhoucke, and
  Alemi]{szegedy2017inception}
Christian Szegedy, Sergey Ioffe, Vincent Vanhoucke, and Alexander~A Alemi.
\newblock Inception-v4, inception-resnet and the impact of residual connections
  on learning.
\newblock In \emph{Proceedings of the AAAI National Conference on Artificial
  Intelligence (AAAI)}, 2017.

\bibitem[Tan \& Le(2019)Tan and Le]{tan2019efficientnet}
Mingxing Tan and Quoc Le.
\newblock Efficientnet: Rethinking model scaling for convolutional neural
  networks.
\newblock In \emph{Proceedings of the International Conference on Machine
  Learning (ICML)}, 2019.

\bibitem[Thrun(1996)]{thrun1996learning}
Sebastian Thrun.
\newblock Is learning the n-th thing any easier than learning the first?
\newblock In \emph{Neurips}, 1996.

\bibitem[Tolstikhin et~al.(2021)Tolstikhin, Houlsby, Kolesnikov, Beyer, Zhai,
  Unterthiner, Yung, Keysers, Uszkoreit, Lucic, and
  Dosovitskiy]{tolstikhin2021mlpmixer}
Ilya Tolstikhin, Neil Houlsby, Alexander Kolesnikov, Lucas Beyer, Xiaohua Zhai,
  Thomas Unterthiner, Jessica Yung, Daniel Keysers, Jakob Uszkoreit, Mario
  Lucic, and Alexey Dosovitskiy.
\newblock Mlp-mixer: An all-mlp architecture for vision, 2021.

\bibitem[Vaswani et~al.(2017)Vaswani, Shazeer, Parmar, Uszkoreit, Jones, Gomez,
  Kaiser, and Polosukhin]{vaswani2017attention}
Ashish Vaswani, Noam Shazeer, Niki Parmar, Jakob Uszkoreit, Llion Jones,
  Aidan~N Gomez, {\L}ukasz Kaiser, and Illia Polosukhin.
\newblock Attention is all you need.
\newblock \emph{Advances in Neural Information Processing Systems (NeurIPS)},
  30, 2017.

\bibitem[Xie et~al.(2017)Xie, Girshick, Doll{\'a}r, Tu, and
  He]{xie2017aggregated}
Saining Xie, Ross Girshick, Piotr Doll{\'a}r, Zhuowen Tu, and Kaiming He.
\newblock Aggregated residual transformations for deep neural networks.
\newblock In \emph{Proceedings of the IEEE International Conference on Computer
  Vision and Pattern Recognition (CVPR)}, 2017.

\bibitem[Xu et~al.(2022)Xu, Pan, Pan, Hoi, Yi, and Xu]{xu2022regnet}
Jing Xu, Yu~Pan, Xinglin Pan, Steven Hoi, Zhang Yi, and Zenglin Xu.
\newblock Regnet: self-regulated network for image classification.
\newblock \emph{IEEE Transactions on Neural Networks and Learning Systems},
  2022.

\bibitem[Yin et~al.(2020)Yin, Molchanov, Alvarez, Li, Mallya, Hoiem, Jha, and
  Kautz]{yin2020dreaming}
Hongxu Yin, Pavlo Molchanov, Jose~M Alvarez, Zhizhong Li, Arun Mallya, Derek
  Hoiem, Niraj~K Jha, and Jan Kautz.
\newblock Dreaming to distill: Data-free knowledge transfer via
  {D}eep{I}nversion.
\newblock In \emph{Proceedings of the IEEE International Conference on Computer
  Vision and Pattern Recognition (CVPR)}, 2020.

\bibitem[Yoon et~al.(2018)Yoon, Yang, Lee, and Hwang]{YoonJ2018iclr}
Jaehong Yoon, Eunho Yang, Jeongtae Lee, and Sung~Ju Hwang.
\newblock Lifelong learning with dynamically expandable networks.
\newblock In \emph{Proceedings of the International Conference on Learning
  Representations (ICLR)}, 2018.

\bibitem[Yoon et~al.(2020)Yoon, Kim, Yang, and Hwang]{yoon2020apd}
Jaehong Yoon, Saehoon Kim, Eunho Yang, and Sung~Ju Hwang.
\newblock Scalable and order-robust continual learning with additive parameter
  decomposition.
\newblock In \emph{Proceedings of the International Conference on Learning
  Representations (ICLR)}, 2020.

\bibitem[Zagoruyko \& Komodakis(2016)Zagoruyko and
  Komodakis]{zagoruyko2016wide}
Sergey Zagoruyko and Nikos Komodakis.
\newblock Wide residual networks.
\newblock \emph{arXiv preprint arXiv:1605.07146}, 2016.

\bibitem[Zenke et~al.(2017)Zenke, Poole, and Ganguli]{zenke17si}
Friedemann Zenke, Ben Poole, and Surya Ganguli.
\newblock Continual learning through synaptic intelligence.
\newblock In \emph{Proceedings of the International Conference on Machine
  Learning (ICML)}, 2017.

\end{thebibliography}
\bibliographystyle{iclr2018_conference}

\newpage
\appendix
\renewcommand\thefigure{\thesection.\arabic{figure}}    
\renewcommand\thetable{\thesection.\arabic{table}}

\paragraph{Organization.} In the supplementary material, we provide the implementation details and hyper-parameter configurations in \cref{sec:experimental_details}. Further, we show the results for class-IL continual learning and additional visualization of examples generated by \textsc{QDI} in \Cref{sec:entire_batches}.

\section{Experimental Details} \label{sec:experimental_details}

We follow the base hyper-parameter setup from \citet{buzzega2020dark} for SCL and HCL experiments.
We use an SGD optimizer for experiments with base learning rate as $0.03$ for all the models. For each new task a total of $200$ epochs are utilized to train the current model and we use the average accuracy metric on the validation set to store the best-model for evaluation for each task. Batch size is set as $32$ and training is conducted on one NVIDIA V100 GPU of $16$G for CIFAR-10 experiment and $32$G for Split CIFAR-100 and Tiny-ImageNet datasets. For knowledge distillation methods, we use $\alpha$ and $\beta$ equal to $1.0$ for Split CIFAR-10, $3.0$ for Split CIFAR-100 and Tiny-ImageNet and $\beta$ equal to $0.3$ with buffer. QDI previous class target labels are evenly sampled across previous tasks. Optimization hyper-parameters for QDI are provided next for three datasets, each shared across all networks during the training run:
\begin{itemize}
    \item {\bf Split CIFAR-10.} Optimization using Adam optimizer of learning rate $0.005$, \[\alpha_{tv}=0.001, \alpha_{\ell_2}=0, \alpha_{\text{feature}}=0.1\] for $0.5$K iterations.$d(\cdot, \cdot)$ is MSE loss.
    \item {\bf Split CIFAR-100.} Optimization using Adam with learning rate $0.03$, \[\alpha_{tv}=0.003, \alpha_{\ell_2}=0.003, \alpha_{\text{feature}}=0.2\] for $0.5$K iterations. $d(\cdot, \cdot)$ is MSE loss.
    \item {\bf Split Tiny-ImageNet.} Optimization based on Adam optimizer of learning rate $0.05$, \[\alpha_{tv}=0.001, \alpha_{\ell_2}=0.05, \alpha_{\text{feature}}=0.5\] for $0.5$K iterations. $d(\cdot, \cdot)$ is MSE loss.
\end{itemize}
For each dataset we found the QDI hyper-parameters based on a validation set obtainend by randomly sampling $10\%$ of the training set. All results in main paper are on the test set, with $3$ independent runs of the hyperset from random seeds for means and standard deviations.

\section{Additional Results}

\noindent{\bf Class-IL continual learning.} \Cref{tab:main_ciltable} shows the results for class-IL continual learning. 
\begin{table*}[t]
\vspace{-0.05in}
\setlength{\tabcolsep}{3pt} %
\resizebox{\textwidth}{!}{
\begin{tabular}{ll@{\hspace{6pt}}ccccccc}
\toprule
& {\textsc{Method}} & $\mathcal{B}$ &\multicolumn{2}{c}{\textsc{Split CIFAR-10}} &\multicolumn{2}{c}{\textsc{Split CIFAR-100}}&\multicolumn{2}{c}{\textsc{Split Tiny-ImageNet}}\\
\midrule
& & & $\mathcal{A}_T~(\uparrow)$ & $\mathcal{F}_T~(\downarrow)$ & $\mathcal{A}_T~(\uparrow)$ & $\mathcal{F}_T~(\downarrow)$ & $\mathcal{A}_T~(\uparrow)$ & $\mathcal{F}_T~(\downarrow)$\\
\midrule
& \multicolumn{8}{c}{\textsc{Standard Continual Learning}} \\
\midrule
& \textsc{Finetune}  & -- & 
{19.60} \scriptsize($\pm$ 0.03) & {96.66} \scriptsize($\pm$ 0.12) &
{6.93} \scriptsize($\pm$ 1.13) & {60.53} \scriptsize($\pm$ 1.11) &
{6.78} \scriptsize($\pm$ 0.14) & {40.67} \scriptsize($\pm$ 0.92) \\

& \textsc{DI}~{\small \citep{yin2020dreaming}}   & -- &  
{22.72} \scriptsize($\pm$ 1.02) &  {77.65} \scriptsize($\pm$ 0.30) &
{7.21} \scriptsize($\pm$ 0.75) &  {63.90} \scriptsize($\pm$ 2.87) &
{7.28} \scriptsize($\pm$ 2.05) &  {51.70} \scriptsize($\pm$ 0.63) \\

& \textsc{SI}~{\small \citep{zenke17si}}$^*$  & -- & 
{19.48} \scriptsize($\pm$ 0.17) &  {95.78} \scriptsize($\pm$ 0.64) &
{--} &  {--} &
{6.58} \scriptsize($\pm$ 0.31) &  {--} \\

& \textsc{LwF}~{\small \citep{LiZ2016learning}}$^*$  & -- &
{19.61} \scriptsize($\pm$ 0.05) &  {96.69} \scriptsize($\pm$ 0.25) &
{--} & {--} &
{8.46} \scriptsize($\pm$ 0.22) &  {--} \\

\cmidrule{2-9}

\rowcolor{lgreen} & \textsc{KD (Ours)}  & -- & 
{\bf 22.73} \scriptsize(\bf $\pm$ 0.75) & {87.00} \scriptsize($\pm$ 2.29) & %
{16.77} \scriptsize($\pm$ 0.63) & {80.65} \scriptsize($\pm$ 0.99) & %
{20.86} \scriptsize($\pm$ 0.14) & {42.38} \scriptsize($\pm$ 0.95) \\

\rowcolor{lgreen} & \textsc{KD w/ QDI (Ours)} & -- &  
{19.75} \scriptsize($\pm$ 0.03) & {\bf 0.08} \scriptsize(\bf $\pm$ 0.02) & %
{\bf 22.53} \scriptsize(\bf $\pm$ 1.15) & {\bf 16.50} \scriptsize(\bf $\pm$ 2.72) & %
{\bf 24.21} \scriptsize(\bf $\pm$ 0.52) & {\bf 36.58} \scriptsize(\bf $\pm$ 1.56) \\

\cmidrule{2-9}

& \textsc{ICARL}~{\small \citep{rebuffi2017icarl}}$^*$  & \checkmark &
{49.02} \scriptsize($\pm$ 3.20) &  {28.72} \scriptsize($\pm$ 0.49) &
{--} & {--} &
{7.53} \scriptsize($\pm$ 0.79) &  {--} \\

& \textsc{A-GEM}~{\small \citep{chaudhry2018efficient}}  & \checkmark & 
{21.98} \scriptsize($\pm$ 0.56) & {92.18} \scriptsize($\pm$ 1.98) &
{5.04} \scriptsize($\pm$ 0.12) & {91.93} \scriptsize($\pm$ 0.22) & 
{7.49} \scriptsize($\pm$ 0.12) & {72.04} \scriptsize($\pm$ 0.34) \\

& \textsc{ER}~{\small ~{\small \citep{rolnick19er}}}  & \checkmark &  
{48.56} \scriptsize($\pm$ 1.40) & {58.11} \scriptsize($\pm$ 1.61) &
{9.65}\scriptsize($\pm$ 0.95) & {85.20} \scriptsize($\pm$ 1.27) &
{10.70} \scriptsize($\pm$ 0.27) &  {83.99} \scriptsize($\pm$ 0.18) \\

& \textsc{DER}~{\small \citep{buzzega2020dark}}  & \checkmark &  
{66.08} \scriptsize($\pm$ 1.18) & {27.40} \scriptsize($\pm$ 2.16) & 
{19.01} \scriptsize($\pm$ 0.74) & {65.06} \scriptsize($\pm$ 0.15) & 
{10.01} \scriptsize($\pm$ 1.52) & {65.66} \scriptsize($\pm$ 3.60) \\

& \textsc{DER++}~{\small \citep{buzzega2020dark}}  & \checkmark &  
{67.23} \scriptsize($\pm$ 1.36) & {26.13} \scriptsize($\pm$ 0.28) & 
{21.38} \scriptsize($\pm$ 0.48) & {56.17} \scriptsize($\pm$ 3.37) & 
{8.23} \scriptsize($\pm$ 0.31) & {68.51} \scriptsize($\pm$ 1.17) \\

\cmidrule{2-9}

\rowcolor{lgreen} & \textsc{KD w/ Buffer (Ours)}  & \checkmark &  
{\bf 70.34} \scriptsize($\pm$ 1.07) & {\bf 16.32} \scriptsize($\pm$ 0.18) & %
{\bf 25.19} \scriptsize(\bf $\pm$ 0.15) & {\bf 40.53} \scriptsize($\pm$ \bf 0.73) & %
{\bf 29.21} \scriptsize(\bf $\pm$ 0.74) & {\bf 16.48} \scriptsize(\bf $\pm$ 0.70) \\

\cmidrule{2-9}
& \textsc{Multitask}$^*$  & -- &  {92.20} \scriptsize($\pm$ 0.15) & \textsc{N/A} & {70.32} \scriptsize($\pm$ 0.48) & \textsc{N/A} & {59.99} \scriptsize($\pm$ 0.19) & \textsc{N/A} \\
\midrule

& \multicolumn{8}{c}{\textsc{Heterogeneous Continual Learning}} \\
\midrule
& \textsc{Finetune}  & -- & 
{21.45} \scriptsize($\pm$ 0.75) & {91.24} \scriptsize($\pm$ 2.33) & 
{5.27} \scriptsize($\pm$ 0.23) & {72.99} \scriptsize($\pm$ 1.94) & 
{7.90} \scriptsize($\pm$ 0.13) & {57.23} \scriptsize($\pm$ 0.11) \\

& \textsc{DI}~{\small \citep{yin2020dreaming}}   & -- &  
{20.67} \scriptsize($\pm$ 1.10) &  {70.97} \scriptsize($\pm$ 2.13) &
{6.20} \scriptsize($\pm$ 1.40) &  {73.70} \scriptsize($\pm$ 1.49) &
{6.39} \scriptsize($\pm$ 0.60) &  {51.14} \scriptsize($\pm$ 0.87) \\

\cmidrule{2-9}

\rowcolor{lgreen} & \textsc{KD (Ours)}  & -- & 
{30.21} \scriptsize($\pm$0.11) & {\bf 27.54} \scriptsize(\bf $\pm$ 1.91) & %
{12.94} \scriptsize($\pm$ 1.13) & {58.89} \scriptsize($\pm$ 1.64) & %
{14.18} \scriptsize($\pm$ 0.35) & {46.91} \scriptsize($\pm$ 0.59) \\

\rowcolor{lgreen} & \textsc{KD w/ QDI (Ours)}  & -- & 
{\bf 33.89} \scriptsize(\bf $\pm$ 3.53) & {31.73} \scriptsize($\pm$ 3.51) & %
{\bf 15.86} \scriptsize($\pm$\bf  1.51) & {\bf 32.73} \scriptsize($\pm$\bf  0.77) & %
{\bf 15.38} \scriptsize($\pm$\bf  1.67) & {\bf 37.27} \scriptsize($\pm$\bf  1.31) \\

\cmidrule{2-9}
& \textsc{ER}~{\small \citep{chaudhry2018efficient}}  & \checkmark &  
{38.77} \scriptsize($\pm$ 1.99) & {69.61} \scriptsize($\pm$ 2.20) &
{7.43} \scriptsize($\pm$ 0.36) & {82.85} \scriptsize($\pm$ 0.20) & 
{7.59} \scriptsize($\pm$ 0.12) & {61.77} \scriptsize($\pm$ 0.31) \\

& \textsc{A-GEM}~{\small \citep{chaudhry2018efficient}}  & \checkmark & 
{19.67} \scriptsize($\pm$ 0.41) & {89.26} \scriptsize($\pm$ 2.89) &
{4.62} \scriptsize($\pm$ 0.02) & {88.78} \scriptsize($\pm$ 0.84) & 
{6.93} \scriptsize($\pm$ 0.11) & {63.93} \scriptsize($\pm$ 0.73) \\

& \textsc{DER}~{\small \citep{buzzega2020dark}}  & \checkmark &  
{44.13} \scriptsize($\pm$ 0.98) & {57.64} \scriptsize($\pm$ 4.76) & %
{10.11} \scriptsize($\pm$ 0.65) & {80.92} \scriptsize($\pm$ 1.35) & %
{8.11} \scriptsize($\pm$ 0.26) & {56.73} \scriptsize($\pm$ 0.81) \\

& \textsc{DER++}~{\small \citep{buzzega2020dark}}  & \checkmark &  
{48.82} \scriptsize($\pm$ 1.75) & {50.11} \scriptsize($\pm$ 2.74) & %
{10.97} \scriptsize($\pm$ 0.55) & {73.62} \scriptsize($\pm$ 0.86) & %
{8.88} \scriptsize($\pm$ 0.61) & {55.54} \scriptsize($\pm$ 2.13)\\

\cmidrule{2-9}

\rowcolor{lgreen} & \textsc{KD w/ Buffer (Ours)}  & \checkmark & 
{\bf 65.40} \scriptsize(\bf $\pm$ 0.96) & {\bf 10.13} \scriptsize(\bf $\pm$ 2.19) & %
{\bf 21.00} \scriptsize(\bf $\pm$ 1.01) & {\bf 38.49} \scriptsize(\bf $\pm$ 1.39) & %
{\bf 18.77} \scriptsize(\bf $\pm$ 0.91) & {\bf 8.76} \scriptsize(\bf $\pm$ 0.79) \\

\bottomrule
\end{tabular}}
\caption{\small {\bf Accuracy and forgetting} with class-IL on standard CL and HCL. The best results are highlighted in {\bf bold}. $\mathcal{B}$ denotes replay-buffer, $\mathcal{A}_T, \mathcal{F}_T$ denote average accuracy and forgetting after the completion of training.\textcolor{black}{$^*$~denotes the methods whose numbers were used from \citet{buzzega2020dark} and $-$ indicates the unavailability of results. All other experiments are over three independent runs.}\label{tab:main_ciltable}}
\end{table*}

\begin{figure*}[b]
\centering

\resizebox{0.83\linewidth}{!}{
\begingroup
\renewcommand*{\arraystretch}{0.3}
\begin{tabular}{cc|cc}
\includegraphics[height=\linewidth,clip,trim=5px 0 0 4px]{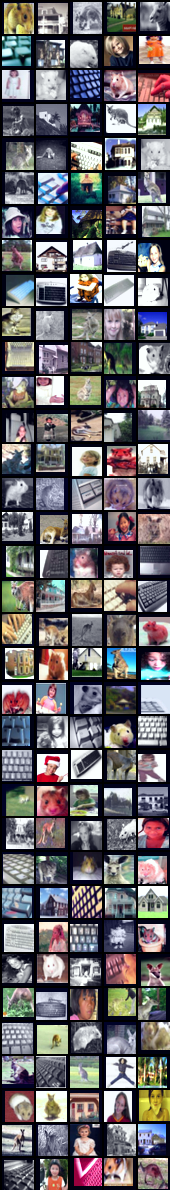} &
\includegraphics[height=\linewidth,clip,trim=5px 0 0 4px]{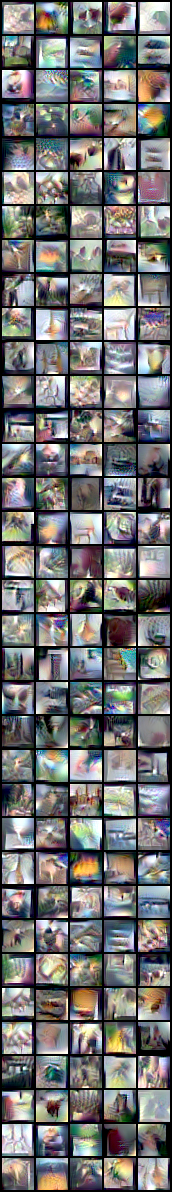} &
\includegraphics[height=\linewidth,clip,trim=5px 0 0 4px]{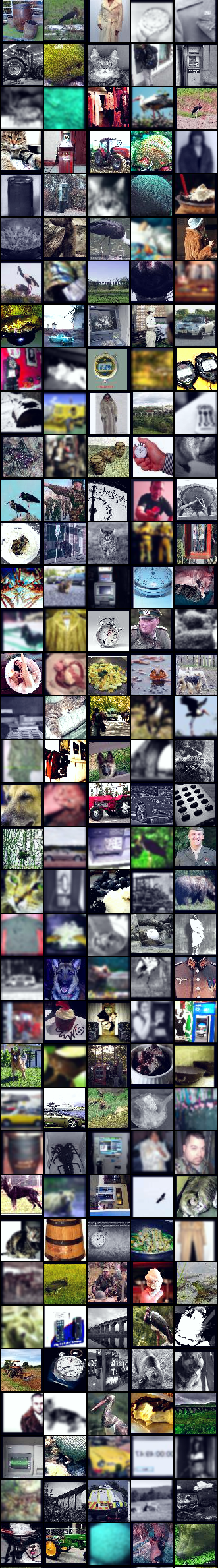} &
\includegraphics[height=\linewidth,clip,trim=5px 0 0 4px]{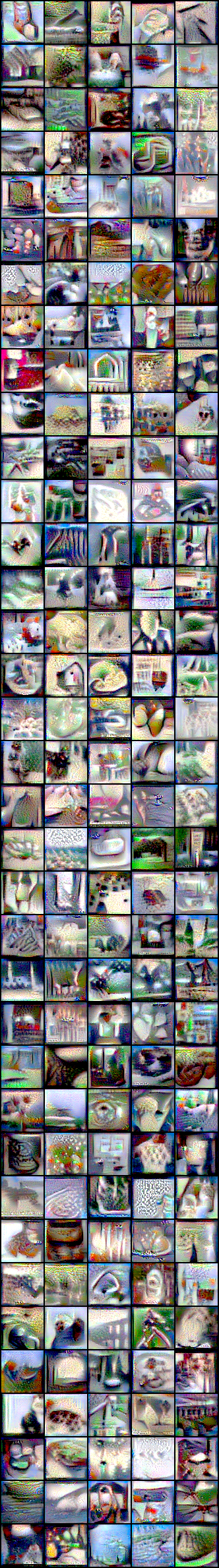} 
\\
\tiny{starting} & \tiny{synthesized} & \tiny{starting} & \tiny{synthesized} \\\\
\multicolumn{2}{c}{\scriptsize{\textbf{CIFAR-100}}} & \multicolumn{2}{c}{\scriptsize{\textbf{Tiny-ImageNet}}}%
\end{tabular}
\endgroup
}

\caption{More QDI visualization for CIFAR-100 and Tiny-ImageNet datasets. Best viewed in color.
}
\label{fig_app:more_samples}
\end{figure*}

\noindent{\bf Visualization of synthesized examples.} \label{sec:entire_batches}
We provide extra visualization of QDI samples for the CIFAR-100 and Tiny-ImageNet datasets as in Fig.~\ref{fig_app:more_samples}. It can be observed that the proposed method scales across datasets with high fidelity in synthesized samples. More interestingly, the optimization step can find very close proxy of the current task semantics in older domains and dream out visual features of high perceptual realism.

\end{document}